# Towards Flexible Teamwork

**Milind Tambe**                                                                TAMBE@ISI.EDU
*Information Sciences Institute and Computer Science Department*
*University of Southern California*
*4676 Admiralty Way*
*Marina del Rey, CA 90292, USA*

## Abstract

Many AI researchers are today striving to build agent teams for complex, dynamic multi-agent domains, with intended applications in arenas such as education, training, entertainment, information integration, and collective robotics. Unfortunately, uncertainties in these complex, dynamic domains obstruct coherent teamwork. In particular, team members often encounter differing, incomplete, and possibly inconsistent views of their environment. Furthermore, team members can unexpectedly fail in fulfilling responsibilities or discover unexpected opportunities. Highly flexible coordination and communication is key in addressing such uncertainties. Simply fitting individual agents with precomputed coordination plans will not do, for their inflexibility can cause severe failures in teamwork, and their domain-specificity hinders reusability.

Our central hypothesis is that the key to such flexibility and reusability is providing agents with general models of teamwork. Agents exploit such models to autonomously reason about coordination and communication, providing requisite flexibility. Furthermore, the models enable reuse across domains, both saving implementation effort and enforcing consistency. This article presents one general, *implemented* model of teamwork, called STEAM. The basic building block of teamwork in STEAM is *joint intentions* (Cohen & Levesque, 1991b); teamwork in STEAM is based on agents' building up a (partial) hierarchy of joint intentions (this hierarchy is seen to parallel Grosz & Kraus's partial Shared-Plans, 1996). Furthermore, in STEAM, team members monitor the team's and individual members' performance, reorganizing the team as necessary. Finally, decision-theoretic communication selectivity in STEAM ensures reduction in communication overheads of teamwork, with appropriate sensitivity to the environmental conditions. This article describes STEAM's application in three different complex domains, and presents detailed empirical results.

## 1. Introduction

> **teamwork:** *cooperative effort by the members of a team to achieve a common goal.* – American Heritage Dictionary

Teamwork is becoming increasingly critical in many multi-agent environments, such as, virtual training (Tambe et al., 1995; Rao et al., 1993), interactive education (for instance, in virtual historical settings, Pimentel & Teixeira, 1994), internet-based information integration (Williamson, Sycara, & Decker, 1996), RoboCup robotic and synthetic soccer (Kitano et al., 1995, 1997), interactive entertainment (Hayes-Roth, Brownston, & Gen, 1995; Reilly, 1996), and potential multi-robotic space missions. Teamwork in such complex, dynamic domains is more than a simple union of simultaneous coordinated activity. An illustrative example provided by Cohen and Levesque (1991b) — worth repeating, given that the difference between simple coordination and teamwork is often unacknowledged in the literature — focuses on the distinction between ordinary traffic and driving in a convoy. Ordinary traffic is simultaneous and coordinated by traffic signs, but it is not considered teamwork. Driving in a convoy, however, is an example of teamwork. The difference in the two situations is that while teamwork does involve coordination, in addition, it at least involves a common team goal and cooperation among team members.

This article focuses on the development of a general model of teamwork to enable a team to act coherently, overcoming the uncertainties of complex, dynamic environments. In particular, in these environments, team members often encounter differing, incomplete and possibly inconsistent views of the world and (mental) state of other agents. To act coherently, team members must flexibly communicate to avoid miscoordination. Furthermore, such environments can often cause particular team members to unexpectedly fail in fulfilling responsibilities, or to discover unexpected opportunities. Teams must thus be capable of monitoring performance, and flexibly reorganizing and reallocating resources to meet any contingencies. Unfortunately, implemented multi-agent systems often fail to provide the necessary flexibility in coordination and communication for coherent teamwork in such domains (Jennings, 1994, 1995). In particular, in these systems, agents are supplied only with preplanned, domain-specific coordination. When faced with the full brunt of uncertainties of complex, dynamic domains, the inflexibility of such preplanned coordination leads to drastic failures — it is simply difficult to anticipate and preplan for all possible contingencies. Furthermore, in scaling up to increasingly complex teamwork situations, these coordination failures continually recur. In addition, since coordination plans are domain specific, they cannot be reused in other domains. Instead, coordination has to be redesigned for each new domain.

The central hypothesis in this article is that providing agents with a general model of teamwork enables them to address such difficulties. Such a model enables agents to autonomously reason about coordination and communication, providing them the requisite flexibility in teamwork. Such general models also allow reuse of teamwork capabilities across domains. Not only does such reuse save implementation effort, but it also ensures consistency in teamwork across applications (Rich & Sidner, 1997). Fortunately, recent theories of teamwork have begun to provide the required models for flexible reasoning about teamwork, e.g., *joint intentions* (Cohen & Levesque, 1991b; Levesque, Cohen, & Nunes, 1990), *SharedPlan* (Grosz, 1996; Grosz & Kraus, 1996; Grosz & Sidner, 1990) and joint responsibility (Jennings, 1995), are some of the prominent ones among these. However, most research efforts have failed to exploit such teamwork theories in building practical applications (Jennings, 1994, 1995).

This article presents an *implemented* general model of teamwork, called STEAM (simply, a **S**hell for **TEAM**work).[1] At its core, STEAM is based on the *joint intentions* theory (Levesque et al., 1990; Cohen & Levesque, 1991b, 1991a); but it also parallels and in some cases borrows from the *SharedPlans* theory (Grosz, 1996; Grosz & Kraus, 1996; Grosz & Sidner, 1990). Thus, while STEAM uses joint intentions as the basic building block of teamwork, as in the SharedPlan theory, team members build up a complex hierarchical structure of joint intentions, individual intentions and beliefs about others' intentions. In STEAM, communication is driven by commitments embodied in the joint intentions theory — team members may communicate to attain mutual belief while building and disbanding joint intentions. Thus, joint intentions provide STEAM a principled framework for reasoning about communication, providing significant flexibility. STEAM also facilitates monitoring of team performance by exploiting explicit representation of team goals and plans. If individuals responsible for particular subtasks fail in fulfilling their responsibilities, or if new tasks are discovered without an appropriate assignment of team members to fulfill them, team reorganization can occur. Such reorganization, as well as recovery from failures in general, is also driven by the team's joint intentions.

---

1. STEAM code (with documentation/traces) is available as an online Appendix.

STEAM's operationalization in complex, real-world domains (described in the next section) has been key in its development to address important teamwork issues discussed above. It has also led STEAM to address some practical issues, not addressed in teamwork theories. One key illustration is in STEAM's detailed attention to communication overheads and risks, which can be significant. STEAM integrates decision theoretic communication selectivity — agents deliberate upon communication necessities vis-a-vis incoherency in teamwork. This decision theoretic framework thus enables improved flexibility in communication in response to unexpected changes in environmental conditions.

Operationalizing general models of teamwork, such as STEAM, necessitates key modifications in the underlying agent architectures. Agent architectures such as Soar (Newell, 1990), RAP (Firby, 1987), PRS (Rao et al., 1993), BB1 (Hayes-Roth et al., 1995), and IRMA (Pollack, 1992) have so far focused on individual agent's flexible behaviors via mechanisms such as commitments and reactive plans. Such architectural mechanisms need to be enhanced for flexible teamwork. In particular, an explicit representation of mutual beliefs, reactive *team* plans and team goals is essential. Additional types of commitments, suitable for a team context, may need to be embodied in the architectures as well. Without such architectural moorings, agents are unable to exploit general models of teamwork, and reason about communication and coordination. This view concurs with Grosz (1996), who states that "capabilities for teamwork cannot be patched on, but must be designed in from the start".

Our operationalization of STEAM is based on enhancements to the Soar architecture (Newell, 1990), plus a set of about 300 domain-independent Soar rules. Three different teams have been developed based on this operationalization of STEAM. These teams have a complex structure of team-subteam hierarchies, and operate in complex environments — in fact, two of them operate in a commercially-developed simulation environment for training. This article presents detailed experimental results from these teams, illustrating the benefits of STEAM in their development.

STEAM is among just a very few implemented general models of teamwork. Other models include Jennings' *joint responsibility* framework in the GRATE* system (Jennings, 1995) (based on Joint Intentions theory), and Rich and Sidner's COLLAGEN (Rich & Sidner, 1997) (based on the SharedPlans theory), that both operate in complex domains. While Section 7 will discuss these in greater detail, STEAM significantly differs from both these frameworks, via its focus on a different (and arguably wider) set of teamwork capabilities that arise in domains with teams of more than two-three agents, with more complex team organizational hierarchies, and with practical emphasis on communication costs.

The rest of the article begins with a concrete motivation for our research via a description of key teamwork problems in real-world domains (Section 2). Section 3 discusses theories of teamwork and sketches their implications for STEAM. Section 4 next describes STEAM, our implemented model of teamwork. Section 5 discusses STEAM's selective communication. Section 6 presents a detailed experimental evaluation. Section 7 discusses related work. Finally, Section 8 presents summary and future work.

## 2. Illustrative Domains and Motivations

This investigation focuses on three separate domains. Two of the domains are based on a real-world distributed, interactive simulator commercially developed for military training (Calder et al., 1993). The simulator enables — via networking of several computers — creation of large-scale, 3D synthetic battlefields, where humans, as well as hundreds or even thousands of intelligent and semi-intelligent agents can co-participate (Tambe et al., 1995).

The first domain, Attack (Figure 1), involves pilot agents for a company of (up to eight) synthetic attack helicopters. The company starts at the home-base, where the commander pilot agent first sends orders and instructions to the company members. The company processes these orders and then begins flying towards their specified *battle position*, i.e., the area from which the company will attack the enemy. While enroute to the battle-position, depending on the orders, the company members may fly together or dynamically split into pre-determined subteams. Once the company reaches a *holding point*, it halts. One or two *scout* helicopters fly forward and first scout the battle position. Based on communication from the scouts, other company members fly forward to the battle position. Here, individual pilots repeatedly mask(hide) their helicopters and unmask to shoot missiles at enemy targets. Once the attack completes, the helicopters regroup and return to their home-base. While enroute to the home-base (or initially towards the battle-position), if any company member spots enemy vehicles posing a threat to the company, it alerts others. The company then evades and bypasses the enemy vehicles, while also protecting itself using guns. When the company returns safely to home-base, it rearms and refuels, readying itself for the next mission. An overview of the overall research and development effort in this domain, simulation infrastructure, milestones, and agent behaviors is presented in (Hill et al., 1997).

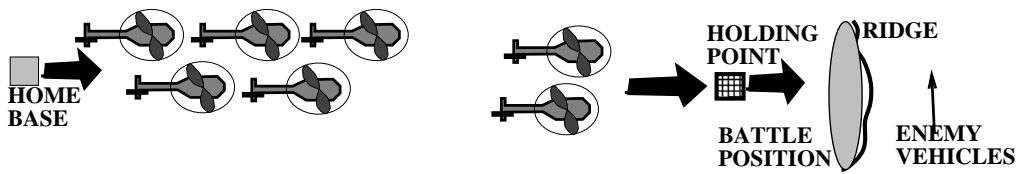

Figure 1: Attack domain: company flying in subteams

In the second domain, Transport (Figure 2), synthetic transport helicopters protected by escort helicopters fly synthetic troops to land. In a typical mission, two or four escort helicopters and four to twelve transport helicopters take off from separate ships at sea to rendezvous at a link-up point. The escorts then provide a protective cover to the transport helicopters during the entire flight to and from their pre-specified landing zone (where the synthetic troops dismount). This domain may involve teams of up to sixteen synthetic pilot agents (the largest team we have encountered); although Figure 2 shows twelve.

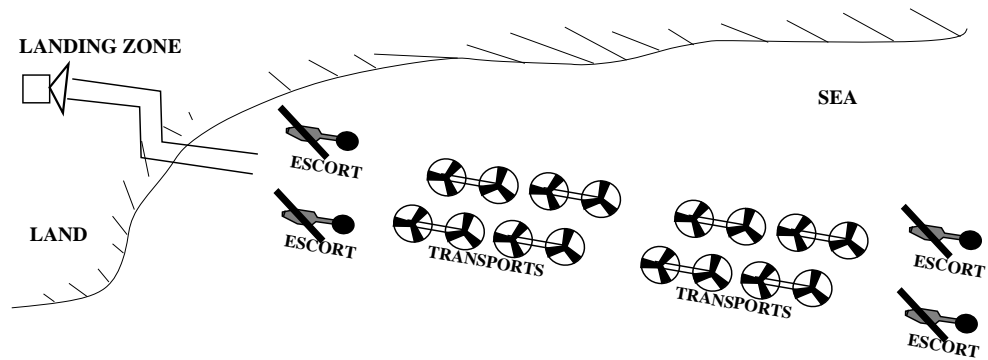

Figure 2: Transport domain with synthetic escort and transport helicopters.

Our third domain is RoboCup synthetic soccer (Kitano et al., 1995). RoboCup is an international soccer tournament for robots and synthetic agents, aimed at promoting research in multi-agent systems. In the synthetic agent track, over 30 teams will participate in the first RoboCup'97 tournament at IJCAI'97 in Japan. The snapshot in Figure 3 shows two competing teams: CMUnited (Stone & Veloso, 1996) versus our *ISI* team.[2]

The Attack domain is illustrative of the teamwork challenges. In our initial, pre-STEAM implementation, the helicopter pilot agents were developed in the Soar integrated agent-architecture (Newell, 1990; Rosenbloom et al., 1991). Each pilot agent was based on a separate copy of Soar. For each such pilot, an operator hierarchy was defined. Figure 4 shows a portion of this hierarchy (Tambe, Schwamb, & Rosenbloom, 1995). Operators are very similar to reactive plans commonly used in other agent architectures, such as the architectures described in Section 1. Each operator consists of (i) precondition rules for selection; (ii) rules for application (a complex operator subgoals); and (iii) rules for termination. At any one point, only one path through this hierarchy is active, i.e., it governs an individual's behavior. For teamwork among individuals, domain-specific coordination plans were added, as commonly done in other such efforts in this type of domain (Rajput & Karr, 1995; Tidhar, Selvestrel, & Heinze, 1995; Laird, Jones, & Nielsen, 1994; Coradeschi, 1997), including our own (Tambe et al., 1995). For instance, after scouting the battle position, a scout executes a plan to inform those waiting behind that the battle position is scouted (not shown in Figure 4).

Initially, with two-three pilot agents and few enemy vehicles, limited behaviors and controlled agent interaction, carefully preplanned coordination was adequate to demonstrate desired behaviors. However, as the numbers of agents and vehicles increased, their behaviors were enriched, and domain experts (human pilots) began to specify complex missions, significant numbers of unanticipated agent interactions surfaced. Faced with the full brunt of the uncertainties in this complex, dynamic environment, the carefully hand-coded, pre-planned coordination led to a variety of teamwork failures in the various demonstrations and exercises in 1995-96. Figure 5 lists a small sample of the teamwork failures, roughly in the order they were encountered.

---

2. Since March 1997, a team of graduate students at the Information Sciences Institute (ISI) has joined in in further research and maintenance of the ISI team. While the author continues to be responsible for the teamwork in the player agents, others have made significant contributions to individual agent behaviors.

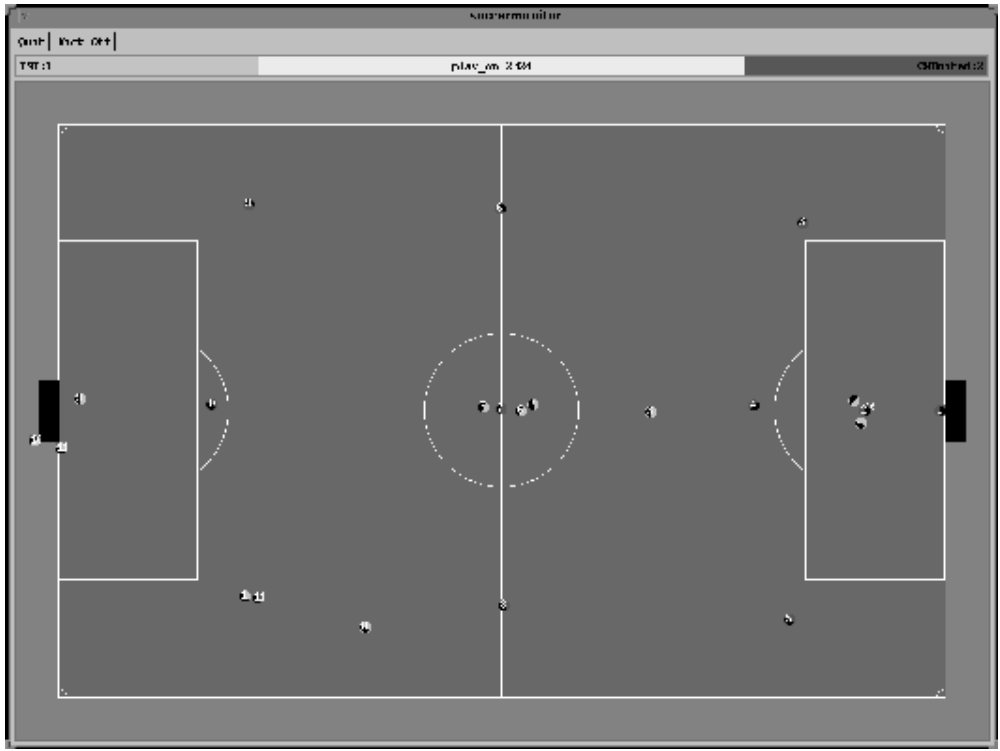

Figure 3: The Robocup synthetic soccer domain.

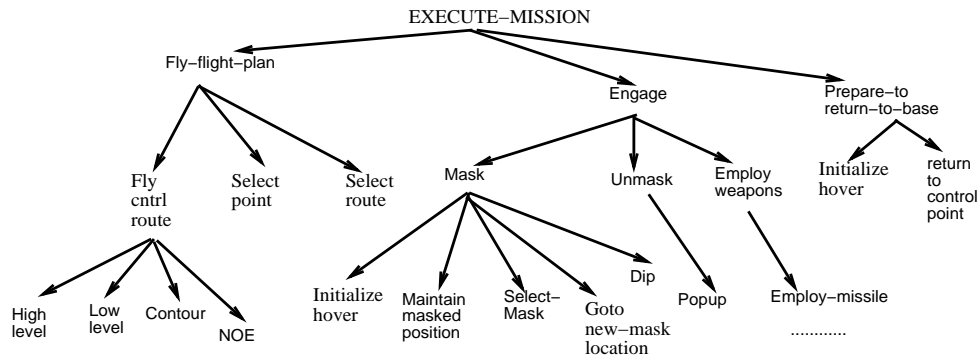

Figure 4: Attack domain: Portion of a pilot agent's operator hierarchy.

One approach to address these failures is a further addition of domain-specific coordination plans; and indeed, this was the first approach we attempted. However, there are several difficulties. First, there is no overarching framework that would enable anticipation of teamwork failures; the teamwork failures just appear to arise unexpectedly. As a result,

1. Upon abnormally terminating engagement with the enemy, the company commander returned to home base alone, abandoning members of its own company at the battle position.

2. Upon reaching the holding area, the company waited, while a single scout started flying forward. Unfortunately, the scout unexpectedly crashed into a hillside; now, the rest of the company just waited indefinitely for the scout's scouting message.

3. One pilot agent unexpectedly processed its initial orders before others. It then flew towards the battle position, while its teammates were left behind at the home base.

4. Only a scout made it to the holding area (all other helicopters crashed or got shot down); but the scout scouted the battle position anyhow, and waited indefinitely for its non-existent company to move forward.

5. When the initial orders unexpectedly failed to allocate the scouting role to team members, the company members waited indefinitely when they reached the holding point.

6. Instructions sent by the commander pilot agent to some company members were lost, because the commander unexpectedly sent them while the members were busy with other tasks. Hence, these members were unable to select appropriate actions.

7. While evading an enemy vehicle encountered enroute, one helicopter pilot agent unexpectedly destroyed the vehicle via gunfire. However, this pilot agent did not inform others; and thus an unnecessary, circuitous bypass route was planned.

8. In an extreme case, when all company members ran out of ammunition, the company failed to infer that their mission could not continue.

9. Two separate companies of helicopters were accidentally allowed to use the same radio channels, leading to interference and loss of an initial message from one of the company commanders — its company hung indefinitely.

Figure 5: Some illustrative examples of breakdown in teamwork.

coordination plans have to be added on a case-by-case basis — a difficult process, since failures have to be first encountered in actual runs. Furthermore, as the system continues to scale up to increasingly complex teamwork scenarios, such failures continually recur. Thus, a large number of special case coordination plans are potentially necessary. Finally, it is difficult to reuse such plans in other domains.

Given these difficulties, we have pursued an alternative approach — provide agents with a general model of teamwork. The agents can then themselves reason about their coordination/communication responsibilities as well as anticipate and avoid (or recover from) teamwork failures. Such an approach also requires an explicit representation of agents' team goals and team plans; for that is the very basis for reasoning about teamwork. Unfortunately, the agent's operator hierarchy shown in Figure 4 represents its own activities. Thus, although the agent is provided information about its teammates, their participation in particular activities is not explicit (but rather, implicit in the coordination plans). As a result, the agent remains ignorant as to which operators truly involve teamwork and the teammates involved in them. For instance, *execute-mission* and *engage* are in reality team activities involving the entire company; while *mask* and *unmask* involve no teamwork. Furthermore, in some team tasks only subteams are involved, adding to the difficulty of relying on implicit representations since the teammates involved in team tasks vary. Even more problematic for implicit representation are team tasks where the team members perform non-identical activities. For instance, consider team tasks such as *travelling overwatch* (where one sub-team travels while the other overwatches), or *wait while battle position scouted* (where scouts scout the battle position while the non-scouts wait). In such tasks, no single agent performs the team activity, and yet it is important to represent and reason about the combined activity that results. This difficulty in representation is not specific to the Soar architecture, but the entire family of architectures mentioned in Section 1.

More importantly, concomitant with the explicit team goals and plans are certain commitments and coordination responsibilities towards the team, based on the general model of teamwork employed. In the absence of both the explicit representation of team goals and plans, as well as commitments and responsibilities they engender, agents are often forced to rely on the problematic domain-specific coordination plans, leading to aforementioned teamwork failures.

## 3. Models of Teamwork

Several teamwork theories have been proposed in the literature (Cohen & Levesque, 1991b; Grosz & Kraus, 1996; Jennings, 1995; Kinny et al., 1992). The theories are not intended to be directly implemented (say via a theorem prover), but to be used as a specification for agent design. They often prescribe general, rather than domain-specific, reasoning processes or heuristics for teamwork. Different types of operational teamwork models could potentially emerge from these theories — the space of such models remains to be fully explored and understood. In developing STEAM, we have focused on the *joint intentions* theory (Cohen & Levesque, 1991b; Levesque et al., 1990; Cohen & Levesque, 1991a), given its detailed formal specification and prescriptive power. The joint intentions theory is briefly reviewed in Section 3.1. STEAM ultimately builds on joint intentions in a way that parallels the *SharedPlan theory* (Grosz & Sidner, 1990; Grosz, 1996; Grosz & Kraus, 1996). The SharedPlans theory is very briefly reviewed in Section 3.2. Section 3.3 sketches the implications of the theories for STEAM. It outlines the rationale for the design decisions in STEAM — in the process, it briefly compares the capabilities provided by the joint intentions and SharedPlan theories. STEAM is later presented in detail in Sections 4 and 5.

### 3.1 Joint Intentions Theory

The joint intentions framework (Cohen & Levesque, 1991b, 1991a; Levesque et al., 1990) focuses on a team's joint mental state, called a *joint intention*. A team $\Theta$ jointly intends a team action if team members are jointly committed to completing that team action, while mutually believing that they were doing it. A joint commitment in turn is defined as a joint persistent goal (JPG). The team $\Theta$'s JPG to achieve $\mathbf{p}$, where $\mathbf{p}$ stands for completion of a team action, is denoted JPG($\Theta$, $\mathbf{p}$, $\mathbf{q}$). $\mathbf{q}$ is an irrelevance clause — as described below, it enables a team to drop the JPG should they mutually believe $\mathbf{q}$ to be false. JPG($\Theta$, $\mathbf{p}$, $\mathbf{q}$) holds iff three conditions are satisfied:

1. All team members mutually believe that $\mathbf{p}$ is currently false.

2. All team members have **p** as their mutual goal, i.e, they mutually know that they want **p** to be eventually true.

3. All team members mutually believe that until **p** is mutually known to be achieved, unachievable or irrelevant they mutually believe that they each hold **p** as a weak goal (WAG).[3] WAG($\mu$, **p**, $\Theta$, **q**), where $\mu$ is a team member in $\Theta$, implies that one of the following holds:

   - $\mu$ believes **p** is currently false and wants it to be eventually true, i.e., **p** is a *normal achievement goal*); or

   - Having privately discovered **p** to be achieved, unachievable or irrelevant (because **q** is false), $\mu$ has committed to having this private belief become $\Theta$'s mutual belief.

JPG provides a basic change in plan expressiveness, since it builds on a team task **p**. Furthermore, a JPG guarantees that team members cannot decommit until **p** is mutually believed to be achieved, unachievable or irrelevant. Basically, JPG($\Theta$, **p**, **q**) requires team members to each hold **p** as a weak achievement goal (WAG). WAG($\mu$, **p**, $\Theta$, **q**), where $\mu$ is a team member in $\Theta$, requires $\mu$ to adopt **p** as its goal if it believes **p** to be false. However, should $\mu$ privately believe that **p** has terminated — i.e., **p** is either achieved, unachievable or irrelevant — JPG($\Theta$,**p**, **q**) is dissolved, but $\mu$ is left with a commitment to have this belief become $\Theta$'s mutual belief. To establish mutual belief, $\mu$ must typically communicate with its teammates about the status of the team task **p**.

The commitment to attain mutual belief in the termination of **p** is a key aspect of a JPG. This commitment ensures that team members stay updated about the status of team activities, and thus do not unnecessarily face risks or waste their time. For instance, consider the first failure presented in Section 5, where the commander returned to home base alone, abandoning its teammates to face a risky situation. Such failures can be avoided given the commitments in a JPG. In our example, the commander would have communicated with its teammates to establish mutual belief about the termination of the engagement.

To enter into a joint commitment (JPG) in the first place, all team members must establish appropriate mutual beliefs and commitments. An explicit exchange of **request** and **confirm** speech acts is one way that a team can achieve appropriate mutual beliefs and commitments (Smith & Cohen, 1996). Since this exchange leads to establishment of a JPG, we will refer to it in the following as the *establish commitments* protocol. The key to this protocol is a persistent weak achievement goal (PWAG). PWAG($\nu i$, **p**, $\Theta$) denotes commitment of a team member $\nu i$ to its team task **p** prior to the team's establishing a JPG.[4] $\mu$ initiates the protocol while its teammates in $\Theta$, $\nu 1$,..,$\nu i$..$\nu n$, respond:

1. $\mu$ executes a **Request**($\mu$, $\Theta$, **p**), cast as an **Attempt**($\mu$, $\phi$, $\psi$). That is, $\mu$'s ultimate goal $\phi$ is to both achieve **p**, and have all $\nu i$ adopt PWAG($\nu i$, **p**, $\Theta$). However, $\mu$ is minimally committed to $\psi$, where $\psi$ denotes achieving mutual belief in $\Theta$ that $\mu$ has the PWAG to achieve $\phi$. With this **Request**, $\mu$ adopts the PWAG.

2. Each $\nu i$ responds via **confirm** or **refuse**. **Confirm**, also an **Attempt**, informs others that $\nu i$ has the PWAG to achieve **p**.

---

3. WAG was originally called WG in (Levesque et al., 1990), but later termed WAG in (Smith & Cohen, 1996).

4. The PWAG also includes an irrelevance clause **q**, but we will not include it here to simplify the following description.

3. If $\forall$ i, $\nu$i confirm, JPG($\Theta$, **p**) is formed.

In establishing a JPG, this protocol synchronizes $\Theta$. In particular, with this protocol, members simultaneously enter into a joint commitment towards a current team activity **p**. While the JPG is the end product of the *establish commitment* protocol, important behavioral constraints are enforced during execution via the PWAGs. In step 1, the adoption of a PWAG implies that if after requesting, $\mu$ privately believes that **p** is achieved, unachievable or irrelevant, it must inform its teammates. Furthermore, if $\mu$ believes that the minimal commitment $\psi$ is not achieved, it must retry (e.g., if a message did not get through it must retransmit the message). Step 2 similarly constrains team members $\nu$i to inform others about **p**, and to rebroadcast. As step 3 indicates, all team members must consent, via confirmation, to the establishment of a JPG. A JPG is not established if any one agent refuses. Negotiations among team members may ensue in such a case; however, that remains an open issue for future work.

## 3.2 Shared Plans Theory

In contrast with joint intentions, the concept of *SharedPlans* (SP) is not based on a joint mental attitude (Grosz, 1996; Grosz & Kraus, 1996; Grosz & Sidner, 1990). Instead, SP relies on a novel intentional attitude, *intending that*, which is similar to an agent's normal *intention to* do an action. However, an individual agent's *intention that* is directed towards its collaborator's actions or towards a group's joint action. *Intention that* is defined via a set of axioms that guide an individual to take actions, including communicative actions, that enable or facilitate its teammates, subteam or team to perform assigned tasks (Grosz & Kraus, 1996).

An SP is either a *full SharedPlan* (FSP) or a *partial SharedPlan*(PSP). We will begin with a definition of an FSP, and then follow with brief remarks about a PSP. An FSP to do $\alpha$ represents a situation where every aspect of a joint activity $\alpha$ is fully determined. This includes mutual belief and agreement in the complete recipe $R_\alpha$ to do $\alpha$. $R_\alpha$ is a specification of a set of actions $\beta_i$, which when executed under specified constraints, constitutes performance of $\alpha$. FSP($P$, **GR**, $\alpha$, $T_p$, $T_\alpha$, $R_\alpha$) denotes a group $GR$'s plan **P** at time $T_p$ to do action $\alpha$ at time $T_\alpha$ using recipe $R_\alpha$. Very briefly, FSP($P$, **GR**, $\alpha$, $T_p$, $T_\alpha$, $R_\alpha$) holds iff the following conditions are satisfied:[5]

1. All members of group $GR$ mutually believe that they each intend that the proposition **Do**(GR, $\alpha$, $T_\alpha$) holds i.e., that $GR$ does $\alpha$ over time $T_\alpha$.

2. All members of $GR$ mutually believe that $R_\alpha$ is the recipe for $\alpha$.

3. For each step $\beta_i$ in $R_\alpha$:

   - A subgroup $GR_k$ ($GR_k \subseteq GR$) has an FSP for $\beta_i$, using recipe $R_{\beta i}$. ($GR_k$ may only be an individual, in which case, it must have a *full individual plan*, an analogue of FSP for individuals.)

   - Other members of $GR$ believe that there exists a recipe such that $GR_k$ can bring about $\beta_i$ and have an FSP for $\beta_i$ (but other members may not know $R_{\beta i}$).

---

5. For the sake of brevity, a context clause $C_\alpha$ is deleted from this definition. Also, in this article, we will not address the *contracting* case discussed in (Grosz & Kraus, 1996).

- Other members of $GR$ intend that $GR_k$ can bring about $\beta_i$ using some recipe.

The SharedPlan theory aspires to describe the entire web of a team's intentions and beliefs when engaged in teamwork. In this endeavor, an FSP represents a limiting case; usually, when engaged in a team activity, a team only has a partial SharedPlan (PSP). The PSP is a snapshot of the team's mental state in a particular situation in their teamwork, and further communication and planning is often used to fulfill the conditions of an FSP (although, in dynamic domains, the team may never actually form an FSP). We focus on three relevant arenas in which partiality may exist in a PSP. First, the recipe $R_\alpha$ may be only partially specified. Certainly, in dynamic environments, such as the ones of interest in our work, recipes could be considered to evolve over time, as teams reactively decide the next step based both on the context and the current situation. For instance, in the Attack domain, the helicopter company may react to enemy vehicles seen enroute, thus evolving their recipe. According to SP theory, team member must arrive at mutual belief in their next step(s) $\beta_i$. For each step $\beta_i$ in the recipe, the relevant subgroup must form a SharedPlan.

Second, the team's task allocation may be *unreconciled*, e.g., the agent or group to perform particular task may not be determined. In this situation, team members intend that there exist some individual or subgroup to do the task. Among actions considered as a result of the intending that, individuals may volunteer to perform the unreconciled task, or persuade/order others to take over the task.

Third, individuals or subgroups may not have attained appropriate mutual beliefs for forming an FSP, leading to communication within the team. Communication may also arise due to agents' "intention that" attitude both towards their team goal and towards teammates' activities. For instance, a team member's intention that its team do an action $\beta_i$, and its belief that communication of some particular information will enable the team to do $\beta_i$, will lead it to communicate that information to the team (as long as such communication does not conflict with previous commitments).

## 3.3 The Influence of Teamwork Theories on STEAM

In STEAM, joint intentions are used as building blocks of teamwork. Several advantages accrue due to this use. First, the commitments in a joint intention begin to provide a principled framework for reasoning about coordination and communication in teamwork. Thus, this framework begins to address teamwork failures such as those in Figure 5. Second, the joint commitments in joint intentions provide guidance for monitoring and maintenance of a team activity, i.e., agents should monitor conditions that cause the team activity to be achieved or unachievable or irrelevant, and maintain the team activity at least until one of these conditions arises. Third, a joint intention leads to an explicit representation of a team activity, and thus facilitates reasoning about teamwork. In particular, as shown later, agents can reason about the relationship between their team activity and an individual's or subteam's contributions to it.

However, a single joint intention for a high-level team goal $\alpha$ is insufficient to provide all of these advantages. To guarantee coherent teamwork, four additional issues must be addressed. Here, the SharedPlans theory helps in analysis of STEAM's approach, and in one case, STEAM directly borrows from SharedPlans. A key observation is that analogous to partial SharedPlans, STEAM builds up snapshots of the team's mental state, but via joint intentions.

The first issue involves coherence in teamwork — team members must pursue a common solution path in service of their joint intention for the high-level team goal $\alpha$. Indeed, as Jennings (1995) observes, without such a constraint, team members could pursue alternative solution paths that cancel each other, so no progress is made towards $\alpha$. The SharedPlan theory addresses such coherence by stepping beyond the team members' "intentions that" towards $\alpha$. In addition, SharedPlans mandates mutual belief in a common recipe (even if partial) and SharedPlans for individual steps $\beta_i$ in the common recipe, thus generating a recursive hierarchy to ensure coherence.

STEAM's approach here parallels that of SharedPlans; however, it builds on joint intentions rather than SharedPlans. That is, STEAM uses joint intentions as a building block to hierarchically build up the mental attitude of individual team members, and ensure that team members pursue a common solution path. In particular, as mentioned earlier, in dynamic domains, given reactive plans, a recipe $R_\alpha$ may evolve step by step during execution. In STEAM, as the recipe evolves, if a step $\beta_i$ requires execution by the entire team, STEAM requires that the entire team agree on $\beta_i$, and form joint intentions to execute it. To execute a substep of $\beta_i$, other joint intentions are formed, leading to a hierarchy. During the expansion of this hierarchy, if a step involves only a subteam then that subteam must form a joint intention to perform that step. If only an individual is involved in the step, it must form an intention to do that step. In general, the resulting intention hierarchy evolves dynamically, depending on the situations the team encounters.

Second, Grosz and Kraus (1996) discuss the tradeoffs in the amount of information team members must maintain about teammates' activities, particularly when a step $\beta_i$ involves only a subteam, or an individual. Grosz and Kraus address this tradeoff in SharedPlans as shown in step 3b in Section 3.2, requiring that team members know only that a recipe exists to enable a teammate(s) to perform its actions, but not the details of the recipe. Similarly, STEAM requires that in case a step $\beta_i$ is performed by a subteam (or just an individual team member), remaining team members track the subteam's joint intention (or the relevant team member's intention) to perform the step. This intention tracking need not involve detailed plan recognition, e.g., as in our previous work (Tambe, 1995, 1996). Instead, a team member must only be able to infer that its teammates intend (or cannot or do not intend) to execute the step $\beta_i$. This minimal constraint is necessary because otherwise, team members may be unable to monitor the current status of the team activity, e.g., that their team activity has fallen apart. In addition, some information about the dependency relationship among team members' actions is useful in monitoring, as discussed in Section 4.2.

A third issue is the analogue of the "unreconciled" case in SharedPlans. STEAM forms a joint intention to replan whenever a team's joint intention for a step $\beta_i$ is seen to be unachievable. Replanning may lead the team to first analyze the cause of the initial unachievability. Among other possibilities, the cause could be the absence of assignment of a subtask to a subteam or individual, or the failure of the relevant individual or subteam in performing the subtask. In such a case, each team member acts to determine the appropriate agent or subteam for performing the relevant task. As a result, an agent can volunteer itself, or suggest to other individuals or subteams to perform the unassigned task.

Of course, the unachievability may be the result of other causes besides lack of assignment; replanning must then address this other cause (further discussion in Section 4.2).

A final issue is generalization of STEAM's communication capabilities via a hybrid approach that combines the prescriptions of the joint intentions approach with some aspects of SharedPlans. A key observation based on (Grosz & Kraus, 1996) is that the communication in joint intentions could potentially be arrived at in SharedPlans via axioms defining *intention that*. For instance, consider that a team member has obtained private information about the achievement of the team's current team action $\beta_1$. In joint intentions, this team member will seek to attain mutual belief in the achievement of $\beta_1$, leading to communication. In contrast, in SharedPlans, the team member's communication would arise because: (i) it *intends that* the team do some action $\beta_2$ which follows $\beta_1$, and (ii) the team cannot do $\beta_2$ without all team members being aware of achievement of $\beta_1$. Thus, further first principles reasoning, based on interrelationships among actions, is required to derive relevant communication in SharedPlans; but in this instance, joint intentions provide for such communication without the reasoning.

In general, if the team's termination of one action $\beta_1$ is essential for the team to perform some following action $\beta_2$, the prescription in joint intentions — to attain mutual belief in termination of team actions — is adequate for relevant communication. However, in some cases, additional communication based on specific information-dependency relationships among actions is also essential. For instance, the scouts in the Attack domain not only inform all company members of completion of their scouting activity (so the company can move forward), but also the precise coordinates of enemy location to enable the company to occupy good attacking positions (information-dependency). Such communication could also be potentially derived from the axioms of *intention that* in SharedPlans, but at the cost of further reasoning.

STEAM does not rely on the first-principles reasoning from intention that for its communication, relying on the prescriptions of joint intentions instead. However, STEAM exploits explicit declaration of information-dependency relationships among actions, for additional communication. Thus, when communicating the termination of a team action $\beta_i$, STEAM checks for any inferred or declared information-dependency relationships with any following action $\beta_j$. The information relevant for $\beta_j$ is also communicated when attaining mutual belief in the termination of $\beta_i$. As a result, based on the specific information-dependency relationship specified, different types of information are communicated, when terminating $\beta_i$. Thus, the scouts can communicate the location of enemy units when communicating the completion of their scouting – given the information-dependency relationship with the planning of attacking positions. If no such relationship is specified, or if other relationships are specified, the scouts would communicate different information.

STEAM thus starts with joint intentions, but then builds up hierarchical structures that parallel the SharedPlans theory, particularly, partial SharedPlans. The result could be considered a hybrid model of teamwork, that borrows from the strengths of both joint intentions (formalization of commitments in building and maintaining joint intentions) and SharedPlans (detailed treatment of team's attitudes in complex tasks, as well as unreconciled tasks). This is of course not the only possible hybrid. As mentioned earlier, further exploration in the space of teamwork models is clearly essential.

## 4. STEAM

STEAM's basis is in executing hierarchical reactive plans, in common with architectures mentioned in Section 1. The novel aspects of STEAM relate to its teamwork capabilities. The key novelty in STEAM is *team operators* (reactive team plans). When agents developed in STEAM select a team operator for execution, they instantiate a *team's joint intentions*. Team operators explicitly express a team's joint activities, unlike the regular "individual operators" which express an agent's own activities. In the hierarchy in Figure 6, operators shown in [] such as [Engage] are team operators, while others are individual operators. Team activities such as *travelling overwatch* or *waiting while battle position scouted* are now easily expressed as team operators, as shown in Figure 6, with activities of individuals or subteams expressed as children of these operators. (Team operators marked with "*" are typically executed by subteams in this domain.)

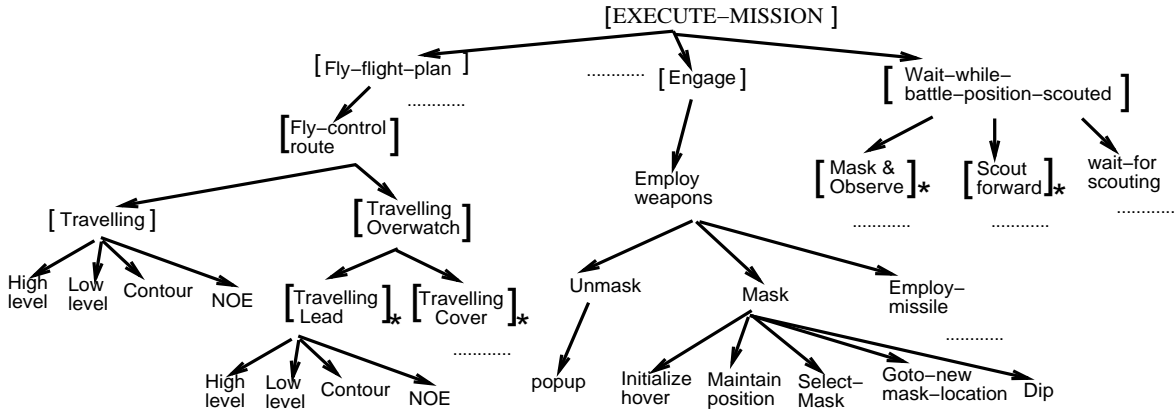

Figure 6: Attack domain: Portion of modified operator hierarchy with *team operators*.

As with individual operators, team operators also consist of: (i) precondition rules; (ii) application rules; and (iii) termination rules. Whether an operator is a team operator or an individual operator is dynamically determined. In particular, when an agent $\nu i$ invokes an operator for execution, the operator is annotated with an "executing agent", which may be dynamically determined to be an individual, or subteam, or a team. If the "executing agent" is a particular team or subteam, the operator is determined to be a team operator. If the "executing agent" is the agent $\nu i$ itself, then an individual operator results. Thus, precise team executing a team operator is not compiled in, but can be flexibly determined at execution time. Figure 6 thus illustrates the configuration of operators that is typical in the Attack domain.

Given an arbitrary team operator OP, all team members must simultaneously select OP to establish a joint intention (joint intention for OP will be denoted as $[OP]_\Theta$). In Figure 6, at the highest level, the team forms a joint intention for $[execute-mission]_\Theta$. In service of this joint intention, the team may form a joint intention $[engage]_\Theta$. In service of $[engage]_\Theta$, individual team members all select individual operators to employ-weapons, thus forming individual intentions. An entire hierarchy of joint and individual intentions is thus formed when an agent participates in teamwork.

A STEAM-based agent maintains its own private state for the application of its individual operators; and a "team state" to apply team operators. A team state is the agent's (abstract) model of the team's mutual beliefs about the world, e.g., in the Transport domain, the team state includes the coordinates of the landing zone. The team state is usually initialized with information about the team, such as the team members in the team, possible subteams, available communication channels for the team, the pre-determined team leader and so forth. STEAM can also maintain subteam states for subteam participation. There is of course no shared memory, and thus each team member maintains its own copy of the team state, and any subteam states for subteams it participates in. To preserve the consistency of a (sub)team state, one key restriction is imposed for modifications to it — only the team operators representing that (sub)team's joint intentions can modify it. Thus, the state corresponding to a subteam $\Omega$ can only be modified in the context of a joint intention $[OP]_\Omega$.

Thus, at minimum, STEAM requires the following modifications to the architectures such as Soar, RAP, PRS and others mentioned in Section 1 to support teamwork: (i) generalization of operators (reactive plans) to represent team operators (reactive team plans); (ii) representation of team and/or subteam states, and (iii) restrictions on team state modifications (only via appropriate team operators). While these team operators and team states are at the foundation of STEAM, as a general model of teamwork, STEAM also involves agents' commitments in teamwork, monitoring and replanning capabilities, and more. Hard-wiring this entire teamwork model within the agent architectures could potentially lead to unnecessary rigidity in agent behaviors. Instead, the STEAM model is maintained as a domain-independent, operational module (e.g., in the form of rules) to guide agents' behaviors in teamwork. In the future, appropriate generalizations of these capabilities could begin to be integrated in agent architectures.

The following subsections now discuss key aspects of STEAM in detail. Section 4.1 discusses team operator execution in STEAM. Section 4.2 describes STEAM's capabilities for monitoring and replanning. Detailed pseudo-code for executing STEAM appears in Appendix A.

## 4.1 Team Operator Execution

To execute a team operator, agents must first establish it as a joint intention. Thus, when a member selects a team operator for execution, it first executes the *establish commitments* protocol described below (introduced in Section 3.1):

1. Team leader broadcasts a message to the team $\Theta$ to establish PWAG to operator OP. Leader now establishes PWAG. If $[OP]_\Theta$ not established within time limit, repeat broadcast.

2. Subordinates $\nu i$ in the team wait until they receive leader's message. Then, turn by turn, broadcast to $\Theta$ establishment of PWAG for OP; and establish PWAG.

3. Wait until $\forall \nu i$, $\nu i$ establish PWAG for OP; establish $[OP]_\Theta$.

With this *establish commitment* protocol, agents avoid problems of the type where just one member flies off to the battle position (item 3, Figure 5). In particular, a team member cannot begin executing the mission without first establishing a joint intention [execute-mission]$_\Theta$. During execution of the *establish commitment* protocol, PWAGs address several contingencies — if an OP is believed achieved, unachievable or irrelevant prior to $[OP]_\Theta$, agents inform teammates. Other contingencies are also addressed, e.g., even if a subordinate initially disagrees with the leader, it will conform to the leader's broadcasted choice of operators. In general, resolving disagreements among team members via negotiation is a significant research problem in its own right (Chu-Carroll & Carberry, 1996), which is not addressed in STEAM. Instead, currently STEAM relies on a team leader to initiate the request, and thus resolve disagreements.

After establishing a joint intention $[OP]_\Theta$, a team operator can only be terminated by updating the team state (mutual beliefs). This restriction on team operator termination avoids critical communication failures of the type where the commander returned to home-base alone — instead, agents must now inform teammates when terminating team operators. Furthermore, with each team operator, multiple termination conditions may be specified, i.e., conditions that make the operator achieved, unachievable or irrelevant. Now, if an agent's private state contains a belief that matches with a team operator's termination condition, and such a belief is absent in its team state, then it creates a communicative goal, i.e., a communication operator. This operator broadcasts the belief to the team, updating the team state, and then terminating the team operator.

As mentioned earlier, during teamwork, an agent may be a participant in several joint intentions, some involving the entire team, some only a subteam. Thus, an agent may be participating in a joint intention involving the entire company, such as $[\text{execute-mission}]_\Theta$, as well as one involving just a subteam, such as $[\text{mask-and-observe}]_\Omega$. When the termination condition of a specific team operator is satisfied, a STEAM-based agent will aim to attain mutual belief in only the relevant subteam or team, e.g., facts relevant to $[\text{mask-and-observe}]_\Omega$ may only be communicated among $\Omega$.

During the broadcast of the communication message, STEAM checks for information-dependency relationships with any following tasks; if one exists, relevant information is extracted from the current world state and broadcast as well. The information-dependency relationship may be specified individually per specific termination condition. For instance, suppose a company member $\nu 4$ sees some enemy tanks on the route while flying to home base. It recognizes that this fact causes the team's current joint intention $[\text{fly-flight-plan}]_\Theta$ to be unachievable. If this fact is absent in the team state, then a communication operator is executed, resulting in a message broadcast indicating termination of the *fly-flight-plan* team operator. In addition, STEAM uses the explicitly specified information-dependency relationship with a following operator *evade* to extract the x,y location and direction of the tank. As a result, the following communication is generated:

> *$\nu 4$ terminate-JPG fly-flight-plan evade tank elaborations 61000 41000 right.*

This message identifies the speaker ($\nu 4$), and informs team members to terminate $[\text{fly-flight-plan}]_\Theta$ in order to evade a tank. Thus, $\nu 4$ informs others; it does not evade tanks on its own. The part of $\nu 4$'s message that follows the key word *elaborations* is due to the information-dependency relationship. This information — the x,y location and direction of the tank — enables team members to evade appropriately. Separating out the information-dependency component in this fashion provides additional communication flexibility, as explained earlier in Section 3.3.[6]

## 4.2 Monitoring and Replanning

One major source of teamwork failures, as outlined in Section 2, is agents' inability to monitor team performance. STEAM facilitates such monitoring by exploiting its explicit representation of team operators. In particular, STEAM allows an explicit specification of monitoring conditions to determine achievement, unachievability or irrelevancy of team operators. In addition, STEAM facilitates explicit specification of the relationship between a team operator and individuals' or subteam's contributions to it. STEAM uses these specifications to infer the achievement or unachievability of a team operator. These specifications are based on the notion of a *role*. A role is an abstract specification of the set of activities an individual or a subteam undertakes in service of the team's overall activity. Thus, a role constrains a team member $\nu i$ (or a subteam $\Omega$) to some suboperator(s) $\text{op}_{\nu i}$ of the team operator $[\text{OP}]_{\Theta}$. For instance, suppose a subteam $\Omega$ is assigned the role of a *scout* in the Attack domain. This role constrains the subteam $\Omega$ to execute the suboperator(s) to scout the battle position in service of the overall team operator *wait-while-battle-position-scouted* (see Figure 6).

Based on the notion of roles, three primitive role-relationships (i) *AND-combination* (ii) *OR-combination* and (iii) *Role-dependency* can currently be specified in STEAM. These primitive role-relationships — called *role-monitoring constraints* — imply the following relationships between a team operator [OP] and its suboperators:

1. *AND-combination*: $[\text{OP}]_{\Theta} \iff \bigwedge_{i=1}^{n} \text{op}_{\nu i}$

2. *OR-combination*: $[\text{OP}]_{\Theta} \iff \bigvee_{i=1}^{n} \text{op}_{\nu i}$

3. *Role dependency*: $\text{op}_{\nu i} \implies \text{op}_{\nu j}$ ($\text{op}_{\nu i}$ dependent on $\text{op}_{\nu j}$)

These primitive role-monitoring constraints may be combined, to specify more complex relationships. For instance, for three agents $\nu i$, $\nu j$ and $\nu k$, with roles $\text{op}_{\nu i}$, $\text{op}_{\nu j}$ and $\text{op}_{\nu k}$, a combination AND-OR role relationship can be specified as $((\text{op}_{\nu i} \bigvee \text{op}_{\nu j}) \bigwedge \text{op}_{\nu k})$. STEAM-based agents can now infer that the role non-performance of $\nu k$ ($\neg \text{op}_{\nu k}$) makes $\text{OP}_{\Theta}$ unachievable; but the role non-performance of just one of $\nu i$ or $\nu j$ is not critical to $\text{OP}_{\Theta}$. Similarly, for two agents $\nu i$ and $\nu j$, both an OR-combination plus role-dependency may be specified as $((\text{op}_{\nu i} \bigvee \text{op}_{\nu j}) \bigwedge (\text{op}_{\nu i} \implies \text{op}_{\nu j}))$. Role monitoring constraints may be specified in terms of individuals' roles, or subteam's roles.

The mechanisms for tracking teammates' role performance or inferring their role non-performance is partly domain dependent. As mentioned in Section 3.3, in some domains, an agent need not know its teammate's detailed plan or track that in detail, but may rely on high-level observations. For instance, in the Attack domain, if a helicopter is destroyed, team members infer role non-performance for the affected team member. In other cases, such as the RoboCup Soccer domain, no such high-level indication is available. Instead, agents need to obtain role performance information via agent tracking (plan recognition) (Tambe, 1995, 1996), e.g., is a player agent in the RoboCup simulation dashing ahead to receive a pass? Communication may be another source of information regarding role non-performance. First, as discussed below, STEAM leads individuals to announce role-changes to the team, and thus other team members indirectly infer role-performance information. Second, as discussed in Section 5.1, STEAM may lead individuals to directly communicate their role non-performance. Additionally, a few domain-independent mechanisms for inferring role performance are provided in STEAM. Thus, role non-performance is inferred if no individual or subteam is specified for performance of a role (as in item 5, Figure 5). Also, if all individuals within a subteam are found incapable of performing their roles, STEAM infers the entire subteam cannot perform its role.

---

6. In the future, to enable STEAM-based agents to communicate with non-STEAM-based agents, a generic communication language may be necessary. While generating natural language is currently outside the scope of STEAM, STEAM does not preclude such a possibility. Alternatively, an artificial communication language, such as (Sidner, 1994) may be used.

If, based on the role-monitoring constraints and the role performance information about teammates, STEAM infers team operator $[OP]_\Theta$ to be unachievable, it invokes $[repair]_\Theta$ for replanning. By casting repair as a team operator, agents automatically ensure the entire team's commitment for their replanning (the entire team is affected if $[OP]_\Theta$ is unachievable). Furthermore, agents inform teammates not only about possible repair results, but also repair unachievability or irrelevancy. The actions taken in service of $[repair]_\Theta$ depend on the context. If $[repair]_\Theta$ was invoked due to $[OP]_\Theta$'s domain-specific unachievability conditions, domain-specific repair is triggered. In contrast, if $[repair]_\Theta$ was invoked due to role-monitoring constraint failures, STEAM leads each agent to first analyze the failure. The analysis may reveal a *critical role failure* — a single role failure causing the unachievability of $[OP]_\Theta$ — which may occur in an AND-combination if any agent or subteam fails in its role; or an OR-combination when all team members are role-dependent on a single individual or a single subteam. For instance, when agents are flying in formation via $[travelling]_\Theta$ (OR-combination), everyone is role-dependent on the lead helicopter. Thus, should the lead crash, a critical role failure occurs.

The action taken in cases of a critical role failure is team reconfiguration, to determine a team member, or subteam, to substitute for the critical role. As mentioned earlier, this situation corresponds to the "unreconciled case" in SharedPlans, discussed in Section 3.2. The steps taken in STEAM in this case are as follows:

1. *Determine candidates for substitution*: Each team member first matches it own capabilities or those of other agents or subteams with the requirements of the critical role. Matching currently relies on domain-specific knowledge. Of course, agents or subteams that are the cause of the critical role failure cannot be candidates for substitution.

2. *Check for critical conflicting commitments*: Once an agent determines possible candidate(s), including itself, it checks for conflicts with candidate's existing commitments to the team. If these commitments are already critical, the candidate is eliminated from consideration. For instance, if the candidate is a participant in a team operator which is an AND-combination, its responsibilities to the team are already critical — even if it possesses relevant capabilities, it cannot take over the role in question. Similarly, the candidate is ruled out if all other team members are role-dependent on it.

3. *Announce role-substitution to the team:* Candidate(s) not ruled out in step 2 can substitute for the role. This could mean an individual volunteering itself, or a team leader volunteering its subteam for the critical role. Since $[repair]_\Theta$ is a team operator, and since role-substitution implies its achievement, any role-substitution is announced to $\Theta$.

4. *Delete non-critical conflicting commitments*: After assuming the new role in the team activity, the relevant individual or subteam members delete their old roles and old commitments.

In the Attack domain, team members can follow the above procedure when recovering from critical role failures such as item 5 in Figure 5. There, since a scouting subteam is not specified, and the relevant operator *wait-while-battle-position-scouted* involves an AND-combination of the scouting role with the non-scouts, a critical role failure occurs. A subteam in the rest of the company is located to possess the capabilities of scouting. The leader of this subteam determines that it can volunteer its subteam for scouting, and announces this change in role to the rest of the team. Members of this subteam then delete conflicting commitments. [wait-while-battle-position-scouted]$_\Theta$ is now executed with this new role assignment. (Since such new role assignments are confined to the local context of individual team operators, and since step 2 explicitly checks for critical conflicts, they do not lead to any global side-effects.)

The entire repair procedure above can invoked in the context of a subteam $\Omega$, rather than the team $\Theta$. In this case, [repair]$_\Omega$ will be invoked as a team operator. STEAM follows an identical repair procedure, in this case enabling individuals or sub-subteams to take over particular critical roles. Furthermore, any repair communication here is automatically restricted within $\Omega$.

In case the failure is a pure *role dependency failure*, only a single dependent agent $\nu i$ is disabled from role performance (because op$_{\nu i} \implies$ op$_{\nu j}$). Here, $\nu i$ must locate another agent $\nu k$ such that op$_{\nu i} \implies$ op$_{\nu k}$. Role dependency failure could involve a subteam $\Omega_i$ instead of an individual; and the subteams engage in an identical repair.

If failure type is *all roles failure*, no agent performs its role; this state is irreparable. In this situation, or in case no substitution is available for a critical role, [repair]$_\Theta$ is itself unachievable. Since the repair of [OP]$_\Theta$ is itself unachievable, a complete failure is assumed, and [complete-failure]$_\Theta$ is now invoked. For instance, in the Attack domain, complete failure implies returning to home base. By casting complete-failure as a team operator, STEAM ensures that team members will not execute such drastic actions without consulting teammates. If only a subteam $\Omega$ or an individual $\nu i$ encounters complete-failure, they infer inability to perform their roles in the team $\Theta$'s on-going activity.

## 5. STEAM: Selective Communication

STEAM agents communicate to establish and terminate team operators. Given the large number of team operators in a dynamic environment, this communication is a very significant overhead (as Section 6 shows empirically), or risk (e.g., in hostile environments). Therefore, STEAM integrates decision-theoretic communication selectivity. Here, STEAM takes into consideration communication costs and benefits, as well as the *likelihood that some relevant information may be already mutually believed*. While this pragmatic approach is a response to the constraints of real-world domains, it is not necessarily a violation of the prescriptions of the joint intentions framework. In particular, the joint intentions framework does not mandate communication, but rather a commitment to attain mutual belief. Via its decision-theoretic communication selectivity, STEAM attempts to follow the most cost-effective method of attaining mutual belief relevant in joint intentions.

Figure 7 presents the decision tree for the decision to communicate a fact F, indicating the termination of [OP]$_\Theta$. Rewards and costs are measured to the team, not an individual. The two possible actions are **NC** (not communicate, cost 0) or **C** (communicate, cost Cc). If the action is **NC**, two outcomes are possible. With probability $(1-\tau)$, F was commonly known anyway, and the team is rewarded $B$ for terminating [OP]$_\Theta$. With probability $\tau$, however, F was not known, and thus there is miscoordination in terminating [OP]$_\Theta$ (e.g., some agents come to know of F only later). Given a penalty $C_{mt}$ for miscoordination, the reward reduces to $B$-$C_{mt}$. If the action is **C**, assuming reliable communication, F is known.

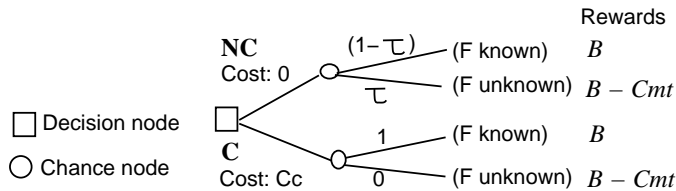

Figure 7: Decision tree for communication.

EU(**C**), the expected utility of option **C**, is $B$-$Cc$ . EU(**NC**) of option **NC** is $B$-$\tau$ *$C_{mt}$. To maximize expected utility, an agent communicates iff EU(**C**) > EU(**NC**), i.e., iff:

$$\tau * C_{mt} > Cc$$

Thus, for instance, in the Attack domain, when flying with high visibility, pilot agents do not inform others of achievement of waypoints on their route, since $\tau$ is low (high likelihood of common knowledge), and $C_{mt}$ is low (low penalty). However, they inform others about enemy tanks on the route, since although $\tau$ is low, $C_{mt}$ is high. The communication cost Cc could vary depending on the situation as well, and team members may flexibly reduce (increase) communication if the cost increases (decreases) during their team activity. Interestingly, if only a single agent is left in a team, $\tau$ drops to zero, and thus, no communication is necessary.

Expected utility maximization is also used for selectivity in the *establish commitments* protocol. If $\gamma$ is the probability of lack of joint commitments, and $C_{me}$ the penalty for executing [OP]$_\Theta$ without joint commitments from the team, then an agent communicates iff EU(**C**) > EU(**NC**), i.e., iff:

$$\gamma * C_{me} > Cc$$

## 5.1 Further Communication Generalization

Further generalization in communication is required to handle uncertainty in the termination criteria for joint intentions. For instance, a team member $\nu4$ may be uncertain that an enemy tank seen enroute causes [fly-flight-plan]$_\Theta$ to be unachievable — the tank's threat may not be clearcut. Yet not communicating could be highly risky. The decision tree for communication is therefore extended to include $\delta$, the uncertainty of an event's threat to the joint intention (Figure 8). Since agents may now erroneously inform teammates to terminate team operators, a nuisance cost -$Cn$ is introduced.

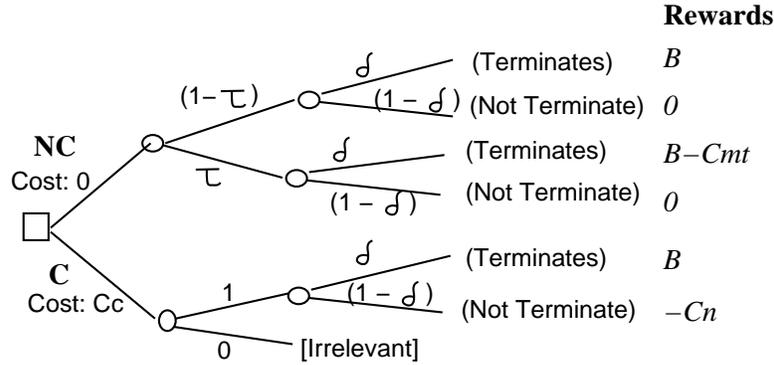

Figure 8: Extended decision tree with $\delta$.

Again, an agent communicates iff EU(**C**) > EU(**NC**), i.e., iff $\delta^* \tau^* C_{mt} > (Cc + (1-\delta)Cn)$. If $\delta$ is 1, i.e., a team operator has terminated, this equation reduces to — $\tau^* C_{mt} > Cc$ — seen previously. If $\delta << 1$, i.e., high uncertainty about termination, no communication results if Cn is high. Therefore, the decision tree is further extended to include a new message type — threat to joint intention — where Cn is zero, but benefits accrued are lower (B - $C_\epsilon$). This threat message maximizes expected utility when $\delta << 1$, i.e., if Cn is high for communicating termination, a team member communicates a threat. For instance, a threat message is used if an agent fails in its own role, which is a threat to the joint intention. However, as before, termination messages are used when $\delta = 1$, where they maximize expected utility.

## 5.2 Estimating Parameters ($\gamma$, $\tau$, $\delta$)

As a first step, STEAM only uses qualitative (low, high, medium) parameter values. STEAM estimates likelihood of lack of joint commitments $\gamma$, via team tracking (Tambe, 1996) — dynamically inferring a team's mental state from observations of team members' actions. Fortunately, rather than tracking each teammate separately, an agent $\nu$i can rely on its own team operator execution for team tracking. In particular, suppose $\nu$i has selected a team operator OP for execution, and it needs to estimate $\gamma$ for operator OP, and its team $\Theta$. Now, if $\nu$i selected OP at random from a choice of equally preferable candidates, then its teammates may differ in this selection. Thus, there is clearly a low likelihood of a joint commitment — $\nu$i estimates $\gamma$ to be high. However, if OP is the only choice available, then $\gamma$ depends on the preceding $[OP2]_\Omega$ that $\nu$i executed with the team $\Omega$. ($\Omega$ may be just a singleton, i.e., OP2 may be an individual operator that $\nu$i executed alone). There are three cases to consider. First, if $\Theta \subset \Omega$ ($\Theta$ is subteam of $\Omega$) or $\Theta = \Omega$, all members of $\Theta$ were jointly executing $[OP2]_\Omega$. Furthermore, $[OP2]_\Omega$ could only be terminated via mutual belief among $\Theta$. Thus, $\Theta$ is likely to be jointly committed to executing the only next choice OP — $\gamma$ is estimated low. Second, if $\Omega \subset \Theta$, some members in $\Theta$ were not jointly participating in team operator execution earlier; hence $\gamma$ is estimated high. Third, if no operator precedes OP, e.g., OP is first in a subgoal, then $\gamma$ is estimated low.

While agents usually infer matching estimates of $\gamma$, sometimes, estimates do mismatch. Therefore, STEAM integrates some error recovery routines. For instance, if an agent $\nu$i estimates $\gamma$ to be low, when others estimate it high, $\nu i$ starts executing the team operator, and only later receives messages for establishing joint commitments. $\nu i$ recovers by stopping current activities and re-establishing commitments. In contrast, if $\nu i$ mis-estimates $\gamma$ to be high, it unnecessarily waits for messages for establishing commitments. STEAM infers such a mis-estimation via reception of unexpected messages; it then conducts a lookahead search to catch up with teammates.

To estimate $\tau$ (the probability that a fact is not common knowledge), STEAM assumes identical sensory capabilities for team members, e.g., if some fact is visible to an agent, then it is also visible to all colocated teammates. However, at present, domain knowledge is also required to model information media such as radio channels, in estimating $\tau$. $\delta$, the probability of an event's threat to a joint intention, is estimated 1 if a fact matches specified termination conditions. Otherwise role monitoring constraints are used, e.g., in an OR-combination, $\delta$ is inversely proportional to the number of team members. The cost parameters, $C_{mt}$, $C_{me}$, and Cc are assumed to be domain knowledge.

## 6. Evaluation

STEAM is currently implemented within Soar via conventions for encoding operators and states, plus a set of 283 rules. Essentially these rules help encode the algorithm in Appendix A; some sample rules are presented in Appendix B. STEAM has been applied in the three domains mentioned earlier: Attack, Transport and RoboCup. Table 1 provides some information about the three domains. Column 1 lists the three domains. Column 2 lists the maximum number of agents per team in each domain. Column 3 shows the possible variations in the sizes of the team. Thus, in the Attack and Transport domains, the team sizes may vary substantially; but not so in RoboCup. Column 4 shows the number of levels in the team organization hierarchy (e.g., the team-subteam-individual hierarchy is a three level hierarchy). Column 5 shows the maximum number of subteams active at any one time.

| Domain name | Max team size | Team size varation | Levels in team hierarchy | Maximum num subteams |
|---|---|---|---|---|
| Attack | 8 | 2-8 | 3 | 2 |
| Transport | 16 | 3-16 | 4 | 5 |
| RoboCup | 11 | 11 | 3 | 4 |

Table 1: The organizational hierarchy in the three domains.

STEAM's application in these three domains provides some evidence of its generality. In particular, not only do these domains differ in the team tasks performed, but as Table 1 illustrates, the domains differ substantially in the team sizes and structure. The rest of this section now uses the three domains in detailed evaluation of STEAM using the criteria of overall performance, reusability, teamwork flexibility, communication efficiency, as well as effort in encoding and modifying teamwork capabilities.

## 6.1 Overall Performance

One key evaluation criterion is the overall agent-team performance in our three domains. Ultimately, STEAM-based agent teams must successfully accomplish their tasks, within their given environments, both efficiently and accurately. This is a difficult challenge in all three domains. Certainly, the Attack and Transport domains involve complex synthetic military exercises with hundreds of other agents. Furthermore, in these domains, the domain experts (expert human pilots) define the pilot teams' missions (tasks), rather than the developers. STEAM-based pilot teams have so far successfully met the challenges in these domains — they have successfully participated in not one, but about 10 such synthetic exercises, where the domain experts have issued favorable written and verbal performance evaluations.

In the RoboCup domain, our player team must compete effectively against teams developed by other researchers worldwide. At the time of writing this article, our player team easily wins against the winner of the pre-RoboCup'96 competition. However, all teams continue to evolve, and researchers continue to field new sophisticated teams. One key test for all the teams in the near future is the RoboCup'97 tournament at the International Joint Conference on Artificial Intelligence (IJCAI), Nagoya, Japan, in August 1997.

## 6.2 Reuse of Teamwork Capabilities

STEAM's inter-domain and intra-domain reusability is approximately measured in Table 2. Column 1 once again lists the three different domains of STEAM's application. Column 2 lists the total number of rules per agent per domain — which include the rules that encode the domain knowledge acquired from domain experts as well as STEAM rules — illustrating complexity of the agents' knowledge base. The number of STEAM rules used in these domains is listed in Column 3. Column 4 measures percent reuse of STEAM rules across domains. (No reuse is shown in STEAM's first domain, Attack). There is 100% reuse in Transport, i.e., no new coordination/communication rules were written — a major saving in encoding this domain. RoboCup, in its initial stages, has lower reuse. Here, due to weakness in spatial reasoning and tracking, agents fail to recognize other team's play, or even own teammates' failures (e.g., in executing a pass), hampering the reuse of rules for role-monitoring constraints, repair and threat detection. With improved spatial reasoning and tracking, reuse may improve in the future. Column 5 lists the total number of team operators specified per domain, illustrating significant intra-domain reuse — essentially, for each team operator, STEAM's entire teamwork capabilities are brought to bear.

| Domain | Total rules | STEAM rules | STEAM reuse | Team operators |
|--------|-------------|-------------|-------------|----------------|
| Attack | 1575 | 283 | first-use | 17 |
| Transport | 1333 | 283 | 100% | 14 |
| RoboCup | 454 | 110 | 38% | 11 |

Table 2: STEAM reusability data.

### 6.3 Flexibility in Teamwork

Teamwork flexibility is closely related with the measures of overall performance and reusability. Since STEAM's entire teamwork capabilities are brought to bear in executing team operators in all of the domains, there are significant improvements in teamwork flexibility. For instance, in benchmark runs of Attack, almost all of the teamwork failures from our earlier implementation are avoided. Certainly, all of the failures in Figure 5 are addressed:

- Items 1 and 7 are addressed because agents must now attain mutual belief in the achievement, unachievability or irrelevancy of team operators. Thus, in item 1, the commander now attains mutual belief that the helicopter company has completed its engagement with the enemy; while in item 7, the irrelevancy of planning a bypass route is communicated to the company.

- Items 3 and 6 are addressed because agents now act jointly by first ensuring the establishment of joint commitments before executing their roles. For instance, a team member does not begin executing the mission as soon as it processes its orders (item 3); rather, it acts jointly with the team, after the team establishes joint commitments to execute the mission.

- Items 2, 4 and 5 are addressed because the team operator *wait-while-battle-position-scouted* is specified to be an AND-combination of the role of the scouts and the non-scouts. Thus, unachievability of team operators is detected, since either the scouts or the non-scouts cannot perform their role, or the scouting-role assignment is unspecified. In items 2 and 4 no repairs are possible, but at the least the company infers a "complete-failure" and returns to home base, instead of waiting indefinitely. In item 5, the unassigned role again leads to unachievability, but repair is possible because one of the remaining subteams can take over the role of the scout.

- Item 8 is addressed since the relevant operator *engage* is now explicitly defined as a team operator with an OR-combination of members' roles. Thus, based on communication from team members, team members' can infer its unachievability.

- Item 9 is addressed because in the establish-commitments protocol, the leader will repeat its message if a response is not heard within time limit. However, in general, attaining mutual belief given the possibility of uncertain communication channels is a notoriously difficult challenge (Halpern & Moses, 1990); and this remains an issue for future work.

As a further illustration of teamwork flexibility in STEAM, we created six variations in the environmental conditions facing the Attack company of helicopter pilots. Each condition required the pilot team to flexibly modify its communication to maintain coherence in teamwork. The six variations are:

1. *Condition 1*: This is the baseline "normal" condition.

2. *Condition 2*: Although similar to condition 1, we assume in addition that certain radio frequencies/channels which were previously separated, are now common. In particular, messages previously assumed to be privately delivered to only the commander agent from its superiors, are now also made available to the other team members.

3. *Condition 3*: Although similar to condition 2, the communication cost is raised from "low" to "medium".

4. *Condition 4*: Although similar to condition 3, we assume in addition that the helicopter team has only a medium priority for ensuring simultaneous attack on the enemy.

5. *Condition 5*: Here, we once again start with the baseline of condition 1, but assume poor visibility in addition. Thus, agents may not accurately estimate their distances.

6. *Condition 6*: In addition to condition 5, here, the company has some flexibility in reaching the battle position. The company is provided with the option of halting at certain key locations, rather than continuing to fly.

The decision-theoretic framework in STEAM enables agents to flexibly respond to the above conditions. Figure 9 plots the number of messages exchanged among team members for each of the six conditions. The total number of messages in three teams — balanced, cautious and reckless — are compared. Balanced agents fully exploit the decision theory framework, and thus illustrate STEAM's flexibility. Cautious agents always communicate, ignoring the decision theory framework. Reckless agents communicate very little (only if high $C_{mt}$, $C_{me}$). Of course, truly reckless agents would likely not communicate at all, so this definition is relaxed here. All three teams work with identical cost models, Cc, $C_{mt}$, and $C_{me}$. The number of agents were fixed in this experiment to four, so all three teams — cautious, balanced and reckless — could be run (as discussed in the next section, it is difficult to run the cautious team with further increase in team size).

Focusing first on the balanced team, it was able to perform its mission under all six conditions, by flexibly decreasing or increasing the number of messages in response. The first set of conditions (conditions 2 through 4) illustrate that the balanced team can reduce its communication in response to the situation faced, e.g., increase in communication cost. However, under conditions 5 and 6, the balanced team can also increase its communication to address the uncertainties. For instance, with condition 5, knowledge of poor visibility automatically leads team members to explicitly communicate achievement of waypoints on their route. In addition, with condition 6, the team has to communicate to establish commitments when deciding to halt or to fly forward.

The cautious team was also able to perform the mission under all six conditions, but it relies on many more messages and remains insensitive to conditions 2-4 that should result in fewer messages. Indeed, its exchange of 10-20 fold more messages than the balanced team to perform an identical task is not only a waste of precious communication resources, but can create risks for the team in hostile environments. (The next subsection will discuss the issue of communication efficiency in more detail.) The reckless team does communicate fewer messages, but it fails to perform its basic mission. Even in the first normal case, this helicopter company gets stuck on the way to the battle position, since a message with medium $C_{mt}$ but high $\tau$ is not communicated. Interestingly, the number of messages increase in the reckless team under conditions 2-4. This is because condition 2 allows the reckless team to avoid getting stuck before reaching the battle position. (In fact, this condition was designed to get the reckless team unstuck.) Since the reckless team can now perform more of the mission — reaching its battle position — more messages are exchanged. Unfortunately, some key messages are still not exchanged, leaving team members stranded in the battle position.

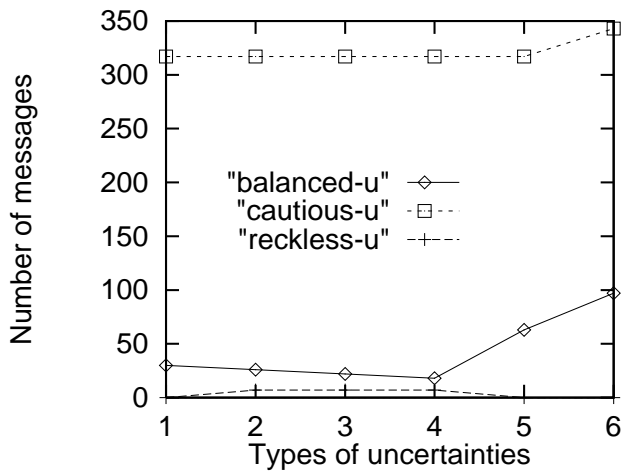

Figure 9: Change in communication with additional uncertainties.

## 6.4 Communication Efficiency

Communication efficiency is critical in teamwork, particularly with scale-up in team size, else communication overheads can cause significant degradation in team performance. Figure 10 and 11 attempt to measure the communication overhead of teamwork, and the usefulness of STEAM's decision-theoretic communication selectivity in lowering the overhead, particularly for a scale-up in team size. Both the figures compare the total number of messages in the three teams introduced above — balanced, cautious and reckless — with increasing numbers of agents per team. In the interest of a fair comparison, the total computational resources available to each team were kept constant (a single SUN Ultra1). While this limits the maximum team size that could be run, the results shown below are sufficiently illustrative in terms of scale-up.

Figure 10 focuses on the Attack domain. Decision-theoretic selectivity enables the balanced team to perform well with few messages — this team is regularly fielded in synthetic exercises. The cautious team exchanges 10 to 20-fold or more messages than the balanced team — a substantial communication overhead. Indeed, beyond six agents, the simulation with cautious team could not be run in real time.[7] Reckless agents in this case do not exchange any messages at all.

Figure 11 focuses on the Transport domain; once again comparing the performance of cautious, balanced and reckless teams for increasing numbers of agents in the team. Once again, decision-theoretic selectivity enables the balanced team to perform well with few messages — this team is regularly fielded in synthetic exercises. The cautious team once again incurs a significant overhead of 10 to 20-fold or more messages than the balanced team. Here, beyond seven agents, the simulation with the cautious team could not be run in real time. Interestingly, in the test scenario for this experiment, the reckless team is able to perform the mission appropriately even though this team exchanges just 1-2 messages, far fewer than the balanced team. To a certain extent, this result illustrates the potential for improving the decision-theoretic selectivity in the balanced team. However, when the test scenario for this experiment was changed, so that the transports arrived late at the rendezvous point, the balanced team was able to continue to perform the mission appropriately. However, the reckless team now performed inappropriately, highlighting the risk in the reckless approach.

---

7. The earlier experiments in Section 6.3 were run with four agents per team, so that the cautious team could be run in real-time.

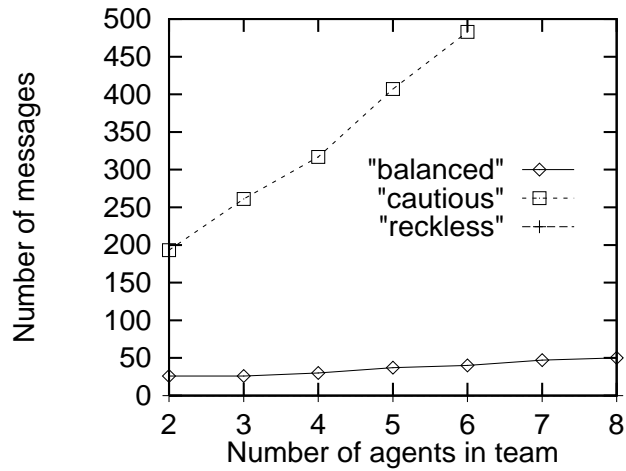

Figure 10: Attack domain: selective communication. Reckless team exchanges no messages and hence that plot overlaps with the x-axis.

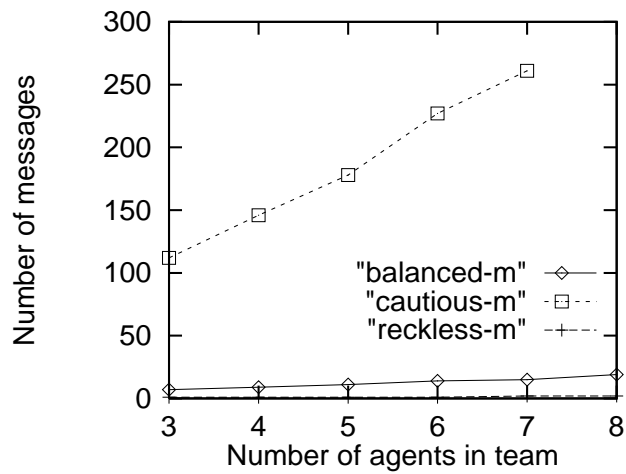

Figure 11: Transport domain: selective communication.

Figure 12 illustrates the differing communication patterns in the cautious and balanced teams for the Attack domain, to attempt to understand the difference in their total com-munication. Figure 12-a plots the varying degree of collaboration (y-axis) during different phases (x-axis) in the Attack domain. Degree of collaboration is measured as the percentage of team operators in a pilot's operator hierarchy (which consists of team and individual operators). A low percentage implies low degree of collaboration and vice versa. The solid line plots the overall degree of collaboration in the team, taking into account all team operators. The dashed line indicates the degree of collaboration without counting team operators executed by this pilot agent's subteam — the differing pattern in the two lines is an indication of the differing degree of subteam activity. In particular, the two lines sometimes overlap but separate out at other times, indicating the flexibility available to the subteam. The overall degree of collaboration is lowest in phases 18-20 (20-40%), where agents engage the enemy. Figure 12-b plots the *percentage* of total communication per phase, for cautious and balanced teams. For instance, the cautious team exchanges 1% of its total messages in phase 20. Communication percentage is correlated to the degree of collaboration per phase for the cautious team (coefficient 0.80), but not for the balanced team (coefficient -0.34). Essentially, unlike the cautious team, the balanced team does not communicate while their collaboration proceeds smoothly.

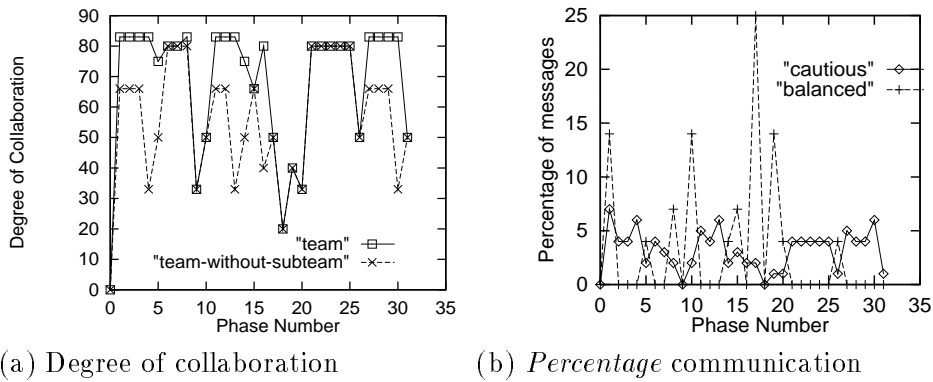

(a) Degree of collaboration    (b) *Percentage* communication

Figure 12: Attack domain: pattern of communication

## 6.5 Encoding and Modification Effort

The final evaluation criteria focus on the effort involved in encoding and modifying agents' teamwork capabilities — comparing the effort in STEAM with alternatives. The key alternative is reproducing all of STEAM's capabilities via special-case coordination plans, as in our initial implementation in the Attack domain. We estimate that such an effort would require significant additional encoding effort. For example, just to reproduce STEAM's selective communication capabilities, our initial implementation could potentially have required hundreds of special case operators. Consider our initial implementation in the Attack domain. Here, the 17 team operators in STEAM (which would only be individual operators in the initial implementation), would each require separate communication operators — two operators each to signal commitments (request and confirm) and one to signal termination of commitments. That is already a total of 51 (17x3). Furthermore, to reproduce selectivity, additional special cases would be necessitated — in the extreme case, each combination of values of ($\tau$, $C_{mt}$, and Cc) or ($\gamma$, $C_{me}$, and Cc), could require a separate special case operator

110

(51 x total combinations, already more than a hundred). Furthermore, separate operators may be required depending on whether the communication occurs with the entire team or only a subteam. Of course, it would appear that all such special cases could be economized in our initial implementation by discovering generalizations — but then STEAM encodes precisely such generalizations to avoid the many special cases.

An additional point of evaluation is easy of modifiability of agent team behaviors. In our experience, domain knowledge acquired from experts is not static — rather it undergoes a slow evolution. In the Attack domain, for instance, real-world military doctrine continues to evolve, requiring modifications in our synthetic pilot team behaviors. In such situations, STEAM appears to facilitate such modifications suggested by domain experts; at least, it is often not necessary to add new coordination plans. For instance, in the Attack domain, domain experts earlier suggested a modification, that the helicopter company should evade enemy vehicles seen enroute, rather than flying over. Here, adding a new unachievability condition for the team operator [fly-flight-plan]$_\Theta$ was sufficient; STEAM then ensured that the pilot agents coordinated the termination of [fly-flight-plan]$_\Theta$, even if just one arbitrary team member detected the enemy vehicles. (Of course, the evasion maneuvers, being domain-specific, had to be added.)

## 7. Related Work

As mentioned earlier, most implementations of multi-agent collaboration continue to rely on domain-specific coordination in service of teamwork (Jennings, 1994, 1995). More recently, however, a few encouraging exceptions have emerged (Jennings, 1995; Rich & Sidner, 1997). We first briefly review these systems and then contrast them with STEAM.

Jennings's (1995) implementation of multi-agent collaboration in the domain of electricity transportation management is also based on joint intentions — it is likely one of the first implementations in a complex domain based on a general model of teamwork. He presents a framework called *joint responsibility* based on a joint commitment to the team's joint goal $\sigma$ and a joint recipe commitment to a common recipe $\Sigma$. Two distinct types of joint commitments — a modification to the joint intentions framework — are claimed necessary because different actions are invoked when joint commitments are dropped. However, as a result, joint responsibility would appear to be limited to a two-level hierarchy of a joint goal and a joint plan, although individuals could execute complex activities in service of the joint plan. The joint responsibility framework is implemented in the GRATE* system, which appears to focus on a team of three agents. In GRATE*, teamwork proceeds with an *organizer* agent detecting the need for joint action; it is then responsible for establishing a team and ensuring members' commitments as required by the joint responsibility method. While the procedure for establishing joint commitments in STEAM is similar to GRATE* — including the similarity of the "leader" in STEAM to the "organizer" in GRATE* — STEAM does benefit from adopting PWAGs, which provides it additional flexibility.

STEAM is also related to COLLAGEN (Rich & Sidner, 1997), a prototype toolkit applied to build a collaborative interface agent for applications such as air-travel arrangements. COLLAGEN's origins are in the SharedPlans theory. Although the COLLAGEN implementation does not explicitly reason from the *intend that* attitude in SharedPlans introduced in (Grosz & Kraus, 1996), it does incorporate discourse generation and interpretation algo-rithms that originate in such reasoning (Lochbaum, 1994). Treating the underlying agent as a blackbox, COLLAGEN facilitates the discourse between a human user and the blackbox (intelligent agent). Several COLLAGEN features aid in such interaction, such as maintenance of a segmented interaction history.

STEAM contrasts with COLLAGEN (Rich & Sidner, 1997) and GRATE* (Jennings, 1995) in several important ways. First, STEAM builds on joint intentions (with some influence of SharedPlans), rather than the SharedPlan approach in COLLAGEN or the joint responsibility approach of GRATE*. Particularly in contrast with joint responsibility, STEAM allows teamwork based on deep joint goal/plan hierarchies. Second, STEAM has the capability for role-monitoring constraints and role substitution in repairing team activities, not relevant in the other two systems. Third, STEAM has attempted scale-up in team size. Thus, STEAM has introduced techniques both to reduce teamwork overheads, e.g., decision-theoretic communication selectivity, as well as to deal with a hierarchy of teams and subteams, not relevant in smaller-scale teams. STEAM also illustrates reuse across domains, not seen in the other two systems. Finally, rather than building a collaboration layer on top of an existing domain-level system or blackbox ("loose coupling"), STEAM has proposed tighter coupling via modifications to support teamwork in the agent architecture itself, e.g., with explicit team goals and team states, and accompanying commitments. The determining factor here would appear to be the tightness of collaboration, e.g., a deeply nested, dynamic joint goal hierarchy should favor a tighter coupling.

In our previous work (Tambe, 1997b) we presented an initial implementation of a teamwork model, also based on joint intentions. That work clearly laid the groundwork for STEAM, by defining team operators, and elaborating on their expressiveness. However, STEAM was later developed because of (i) several problems in that work in continued development of teamwork capabilities in the Attack domain, (ii) the presence of new domains such as Transport, and (iii) significant scale-up in team sizes. Since STEAM both extends and substantially revises that earlier work, it is best to treat STEAM as a separate system, rather than an extension of that early work. STEAM also provides a conceptual advance in a clearer analysis and specification of the joint mental attitude it builds up in a team. In particular, via an explicit analogy to partial SharedPlans (Grosz & Kraus, 1996), this article has spelled out the requirement for teams and subteams to build up a hierarchy of joint intentions, beliefs about other team members' intentions, and joint intentions for the "unreconciled case". This analysis also led to a generalization of communication based on information-dependency.

The following now presents a detailed comparison between STEAM and the earlier work (Tambe, 1997b) in terms of their capabilities. To begin with, STEAM includes an explicit mechanism to establish joint commitments based on PWAGs, which was unaddressed in previous work — so earlier, agents would implicitly, and hence sometimes incorrectly, assume the existence of joint commitments. Also, in earlier work, monitoring and repair was highly specialized. In particular, the mechanism provided for monitoring was based on comparing achievement conditions of operators; this mechanism was later discovered to be limited to monitoring and repair of just one pre-determined specialist role per team operator. Furthermore, the role-substitution was defined via a special procedure executed separately by individuals. In contrast, STEAM has significantly generalized monitoring and repair via its explicit role-monitoring constraints, that enable monitoring of a much greater variety of failures, e.g., the "specialist" is just one case in all of the varied role-monitoring constraint combinations. Furthermore, STEAM establishes a joint intention to resolve all failures, rather than relying on any special case procedures. This is not merely a conceptual advance in terms of an integrated treatment of repair, but has real behavioral implications in providing additional flexibility embodied in the commitments in [repair]$_\Theta$. Furthermore, STEAM's repair generalizes to subteams, addresses previous critical commitments, as well as unallocated tasks. In terms of practical concerns, our previous work (Tambe, 1997b) raised the issue of communication risk in hostile environments, but suggested only a heuristic evaluation of communication costs and benefits; a general purpose mechanism was lacking. STEAM has filled the gap with its decision theoretic framework that now considers various uncertainties, both for selective communication as well as enhancements in communication. Also, unlike STEAM, our earlier work did not deal with complex team-subteam hierarchies, and its mechanisms did not generalize to subteams. Finally, STEAM is backed up with detailed experimental results about *both* its flexibility and reuse across domains, all outside the scope of the previous work.

STEAM is also related to coordination frameworks such as *Partial Global Planning*(PGP) (Durfee & Lesser, 1991), and *Generalized Partial Global Planning*(GPGP)(Decker & Lesser, 1995). Although not driven via theories of collaboration, these coordination frameworks also strive towards domain independence. The earlier work on PGP focuses on a system of cooperating agents for consistent interpretation of data from a distributed sensor network (Durfee & Lesser, 1991). Here, subordinate agents may exchange their individual goals and plans of action. An assigned agent (e.g., a supervisor) may recognize that individual plans of different agents meld into a partial global plan (PGP) — so called because PGPs involve more than one agent but not necessarily all agents (partially global) — in service of a common group goal. The PGP is a basis for planning coordination actions; and it may be transmitted to subordinates for guidance in execution of individual actions. (PGP can accommodate different types of organizations as well.) GPGP (Decker & Lesser, 1995) provides several independent coordination modules, any subset of which may be combined in response to coordination needs of a task environment; the GPGP approach can duplicate and extend the PGP algorithm.

As a general model of teamwork, STEAM can provide a principled underlying model to reason about at least some of the coordination specified in PGP, e.g., agents would establish a joint intention towards the collective goal in a PGP, and modulate their communication via decision-theoretic reasoning. That is, PGP "compiles out" some of the underlying reasoning in STEAM, and thus STEAM could provide additional flexibility in coordination. Essentially, PGP and GPGP do not separate out coordination in teamwork from coordination in general (such as via a centralized coordinator). As a result, they fail to exploit the responsibilities and commitments of teamwork in building up coordination relationships. Analogously, some of the general coordination in PGP or GPGP is unaccounted for in STEAM, and hence understanding relationships among STEAM and GPGP is an interesting area of future work. There is a similar relationship between STEAM and the COOL coordination framework (Barbuceanu & Fox, 1996). COOL also focuses on general purpose coordination by relying on notions of obligations among agents. However, it explicitly rejects the notion of joint goals and joint commitments. It would appear that individual commitments in COOL would be inadequate in addressing some teamwork phenomena, but further work is necessary in understanding the relationship among COOL and STEAM.

In team tracking (Tambe, 1996), i.e., inferring team's joint intentions, the expressiveness of team operators has been exploited. However, issues of establishing joint commitments, communication, monitoring and repair are not addressed. The formal approach to teamwork in (Sonenberg et. al., 1994) transforms team plans into separate role-plans for execution by individuals, with rigidly embedded communications. STEAM purposely avoids such transformations, so agents can flexibly reason with (i) explicit team goals/plans; and (ii) selective communication (seen to be important in practice). In (Gmytrasiewicz, Durfee, & Wehe, 1991), decision theory is applied for message prioritization in coordination based on the agents' recursive modeling of each others' actions. STEAM applies decision theory for communication selectivity and enhancements, but in a very different context — practical operationalization of general, domain-independent teamwork model based on joint intentions.

## 8. Summary and Future Work

Teamwork is becoming increasingly critical in a variety of multi-agent environments, ranging from virtual environments for training and education, to internet-based information integration, to potential multi-robotic space missions (Tambe et al., 1995; Rao et al., 1993; Pimentel & Teixeira, 1994; Williamson et al., 1996; Kitano et al., 1997; Hayes-Roth et al., 1995; Reilly, 1996). In previous implementations of multi-agent systems, including our own, teamwork has often been based on pre-defined, domain-specific plans for coordination. Unfortunately, these plans are inflexible and thus no match for the uncertainties of complex, dynamic environments. As a result, agents' coherent teamwork can quickly dissolve into miscoordinated misbehavior. Furthermore, the coordination plans cannot be reused in other domains. Such reuse is important however, both to save implementation effort and enforce consistency across applications.

Motivated by the critical need for teamwork flexibility and reusability, this article has presented STEAM, a general model of teamwork. While STEAM's development is driven by practical needs of teamwork applications, its core is based on principled theories of teamwork. STEAM is one of just a few implemented systems that have begun to bridge the gap between collaboration theories and practice. STEAM combines several key novel features: (i) use of joint intentions as a building block for a team's joint mental attitude (Levesque et al., 1990; Cohen & Levesque, 1991b) — the article illustrates that STEAM builds up a hierarchical structure of joint intentions and individual intentions, analogous to the partial SharedPlans (Grosz & Kraus, 1996); (ii) integration of novel techniques for explicit establishment of joint intentions (Smith & Cohen, 1996); (iii) principled communication based on commitments in joint intentions; (iv) use of explicit role-monitoring constraints as well as repair methods based on joint intentions; (v) application of decision-theoretic techniques for communication selectivity and enhancements, within the context of the joint intentions framework. To avail of the power of a model such as STEAM, a fundamental change in agent architectures is essential — architectures must provide explicit support for representation of and reasoning with team goals, (reactive) team plans and team states. STEAM has been applied and evaluated in three complex domains. Two of the domains,

Attack and Transport, are based on a real-world simulation environment for training, and here our pilot agent teams have participated large-scale synthetic exercises with hundreds of other synthetic agents. In the third domain, RoboCup, our player agent team is now under development for participation in the forthcoming series of (simulated) soccer tournaments, beginning at IJCAI-97.

Of course, STEAM is far from a complete model of teamwork, and several major issues remain open for future work. One key issue is investigating STEAM's interactions with learning. Initial experiments with chunking (Newell, 1990) (a form of explanation-based learning (Mitchell, Keller, & Kedar-Cabelli, 1986)) in STEAM reveal that agents could automatize routine teamwork activities, rather than always reasoning about them. Specifically, from STEAM's domain-independent reasoning about teamwork, agents learn situation-specific coordination rules. For instance, when the formation leader crashes, another agent learns situation-specific rules to take over as formation lead and communicate. A well-practiced team member could thus mostly rely on learned rules for "routine" activities, but fall back on STEAM rules if it encounters any unanticipated situations. Additionally, STEAM's knowledge-intensive to learning approach could complement current inductive learning approaches for multi-agent coordination (Sen, 1996).

Failure detection and recovery is also a key topic for future work, particularly in environments with unreliable communication. One novel approach exploits agent tracking (Tambe & Rosenbloom, 1995; Tambe, 1996) to infer teammates' high-level goals and intentions for comparison with own goals and intentions. Differences in goals and intentions may indicate coordination failures, since teammates often carry out identical or related tasks. However, given the overheads of such an approach, it has to be carefully balanced with an agents' other routine activities. Initial results of this approach are reported in (Kaminka & Tambe, 1997).

Enriching STEAM's communication capabilities in a principled fashion is yet another key topic for future work. Such enriched communication may form the basis of multi-agent collaborative negotiation (Chu-Carroll & Carberry, 1996). Currently, STEAM relies on the team or subteam leader when resolving disagreements, particularly when deciding the next action. While leadership in teamwork is by itself an interesting phenomena of investigation, enabling agents to negotiate their plans without a leader would also improve STEAM's flexibility. We hope that addressing such issues would ultimately lead STEAM towards improved flexibility in teamwork.

## Acknowledgements

This research was supported as part of contract N66001-95-C-6013 from ARPA/ISO. This article is an extended version of a previous conference paper (Tambe, 1997a). I thank Johnny Chen, Jon Gratch, Randy Hill and Paul Rosenbloom for their comments and support for the work reported in this article. Discussions with Nick Jennings have helped improve the quality of the article. I also thank ISI team members working on the RoboCup effort for their support of the work reported in this article. Domain expertise for this work was provided by David Sullivan and Greg Jackson of BMH Inc., and Wayne Sumner of RDA Logicon.

## Appendix A: Detailed STEAM Specification

The pseudo-code described below follows the description of STEAM provided in this article. It is based on execution of hierarchical operators, or reactive plans. All operators in the hierarchy execute in parallel, and hence the "in parallel" construct. The comments in the pseudo code are enclosed in /* */. The terminology is first described below, to clarify the pseudo-code.

- *Execute-Team-Operator*($\alpha$, $\Theta$, $\mathbf{C}$, $\{\rho 1, \rho 2,...,\rho n\}$) denotes the execution of a team operator $\alpha$, by a team $\Theta$, given the context of the current intention hierarchy $\mathbf{C}$, and with parameters $\rho 1$, $\rho 2...\rho n$.

- Terms $\gamma$, $C_{me}$, Cc, $\tau$, $C_{mt}$ are all exactly as in Section 5.

- $[\alpha]_\Theta$ denotes the team $\Theta$'s joint intention to execute $\alpha$.

- **status**($[\alpha]_\Theta$, *STATUS-OF-*$\alpha$) denotes the status of the joint intention $[\alpha]_\Theta$, whether it is mutually believed to be achieved, unachievable or irrelevant.

- **satisfies** (Achievement-conditions($\alpha$), *f*) denotes that the fact *f* satisfies the achievement conditions of the team operator $\alpha$; similarly with respect to unachievability and irrelevancy conditions.

- *Communicate*(terminate-jpg($\alpha$), *f*,$\Theta$) denotes communication to the team $\Theta$ to terminate $\Theta$'s joint commitment to $\alpha$, due to the fact f.

- Update-state (**team-state**($\Theta$), *f*) denotes the updating of the team state of $\Theta$ with the fact f.

- Update-status($[\alpha]_\Theta$) denotes the updating of the team operator $\alpha$ with its current status of achievement, unachievability or irrelevancy.

- **Agent**($\alpha$) is the individual agent or team executing operator $\alpha$.

- **actions**($\alpha$) denote the actions of the operator $\alpha$.

- **teamtype**($\psi$) is a test of whether the agent $\psi$ is a team or just one individual.

- **self**($\psi$) is a test of whether the agent $\psi$ denotes self.

- **agent-status-change**($\mu$) denotes change in the role performance capability of agent or subteam $\mu$.

- *Execute-individual-Operator*($\psi$, self, $\mathbf{C}$, $\{\rho 1, \rho 2,...,\rho n\}$) denotes the execution of an individual operator $\psi$ by self, given the context of the current intention hierarchy $\mathbf{C}$, and with parameters $\rho 1$, $\rho 2...\rho n$.

For expository purposes, "Execute-team-operator" and "Execute-individual-operator" are defined as separate procedures. In reality, STEAM does not differentiate between the two.

## Team Operator Execution

*Execute-Team-Operator*($\alpha$, $\Theta$, **C**, $\{\rho 1, \rho 2,...,\rho n\}$)
{

1. estimate $\gamma$; /* See Section 5. */

2. if $\gamma$* $C_{me}$ > Cc execute *establish commitments* protocol;
   /* see Section 4.1 for explanation. */

3. establish joint-intention $[\alpha]_\Theta$;

4. While NOT(**status**($[\alpha]_\Theta$, **Achieved**) $\bigvee$ **status**($[\alpha]_\Theta$, **Unachievable**) $\bigvee$ **status**($[\alpha]_\Theta$, **Irrelevant**)) Do
   {

   (a) if (**satisfies** (Achievement-conditions($\alpha$), f) $\bigvee$ **satisfies** (Unachievability-conditions($\alpha$), f) $\bigvee$ **satisfies** (Irrelevance-conditions($\alpha$), f))
   /* This is the case where fact f is found to satisfy the termination condition of $\alpha$. The case where f is only a threat to $\alpha$ (see Section 5.1) is analogous. */

   {
   i. estimate $\tau$; /* see section 5. */
   ii. if $\tau$* $C_{mt}$ > Cc propose-operator *Communicate*(terminate-jpg($\alpha$), f, $\Theta$) with high priority;
   /* See Section 5 and 4.1*/
   iii. if no other higher priority operator, in parallel
   *Execute-individual-operator*(*Communicate*(terminate-jpg($\alpha$), f, $\Theta$), *self*, $\alpha$/**C**, $\{\rho 1, \rho 2,...\}$);
   iv. Update-state (**team-state**($\Theta$), f);
   v. Update-status($[\alpha]_\Theta$);
   }

   (b) if **agent-status-change**($\mu$), where $\mu \in \Theta$
   {
   i. Evaluate role-monitoring constraints; /* See Section 4.2. */
   ii. if role-monitoring constraint failure **cf** such that (**satisfies** (Unachievability-conditions($\alpha$), **cf**) then update-status($[\alpha]_\Theta$);

   }

   (c) if receive communication of terminate-jpg($\alpha$) and fact f
   {
   if (**satisfies** (Achievement-conditions($\alpha$), f) $\bigvee$ **satisfies** (Unachievability-conditions($\alpha$), f) $\bigvee$ **satisfies** (Irrelevance-conditions($\alpha$), f))
   {
   i. Update-state (**team-state**($\Theta$), f);
   ii. Update-status($[\alpha]_\Theta$);
   }
   }

   (d) Update-state(**team-state**($\Theta$), **actions**($\alpha$));
   /* execute domain-specific actions to modify team state of $\Theta$ */

(e) if children operator $\beta 1, \beta 2, \ldots \beta n$ of $\alpha$ proposed as candidates

{

 i. $\beta i \leftarrow$ select-best$\{\beta 1 \ldots \beta n\}$;

 ii. if (**teamtype**($\mathbf{Agent}(\beta i)$) $\bigwedge$ ($\Theta = \mathbf{Agent}(\beta i)$)) then in parallel
  *Execute-team-operator*($\beta i$, $\Theta$, $\alpha/\mathbf{C}$, $\{\rho 1, \rho 2 \ldots\}$);

 iii. if (**teamtype**($\mathbf{Agent}(\beta i)$) $\bigwedge$ ($\mathbf{Agent}(\beta i) \subset \Theta$)) then in parallel

  {

  A. *Execute-team-operator*($\beta i$, $\mathbf{Agent}(\beta i)$, $\alpha/\mathbf{C}$, $\{\rho 1, \rho 2 \ldots\}$);

  B. Instantiate role-monitoring constraints;

  }

 iv. if **self**($\mathbf{Agent}(\beta i)$) then in parallel

  {

  A. *Execute-individual-operator*($\beta i$, *self*, $\alpha/\mathbf{C}$, $\rho 1 \ldots$);

  B. Instantiate role-monitoring constraints;

  }

}

} /* End while statement in 4 */

5. terminate joint intention $[\alpha]_\Theta$;

6. if **status**($[\alpha]_\Theta$, **Unachievable**)

{

if ($\alpha \mathrel{!=} Repair$) /* If $\alpha$ is not itself Repair */

{

*Execute-team-operator*($Repair$, $\Theta$, $\mathbf{C}$, $\{\alpha$, cause-of-unachievability,...$\}$)

/* Repair is explained in detail in Section 4.2. Cause-of-unachievability, passed as a parameter
to Repair, may be role-monitoring constraint violation as in case 4b, or the domain-specific
unachievability conditions. */

} else {

*Execute-team-operator*($Complete$-$Failure$, $\Theta$, $\mathbf{C}$, $\{\alpha$, cause-of-unachievability,...$\}$)

/* If Repair is itself unachievable, complete-failure results, as in Section 4.2 */

}

}

} /* end procedure execute-team-operator */

## Individual Operator Execution

*Execute-individual-Operator*($\psi$, self, $\mathbf{C}$, $\{\rho 1, \rho 2, \ldots, \rho n\}$)

{

1. establish $\psi$ as an individual intention;

2. While NOT(**status**($\psi$, **Achieved**) $\bigvee$ **status**($\psi$, **Unachievable**) $\bigvee$ **status**($\psi$, **Irrelevant**))
Do

{

(a) if (**satisfies** (Achievement-conditions($\psi$), $f$) $\bigvee$ **satisfies** (Unachievability-conditions($\psi$),
$f$) $\bigvee$ **satisfies** (Irrelevance-conditions($\psi$), $f$))

{

 i. Update-state (**state**(self), $f$);

       ii. Update-status($\psi$);

      }

  (b) Update-state(**state**(self), **actions**($\psi$));
     /* execute domain-specific actions to modify private state */

  (c) if new children operator $\{\beta 1...\beta n\}$ of $\psi$ proposed
     {

      i. $\beta$i $\leftarrow$ select-best$\{\beta 1...\beta n\}$;

      ii. *Execute-individual-operator*($\beta$i, *self*, $\psi$/**C**, $\{\rho 1...\}$)

     }

  } /* end while statement in 2 */

3. if **status**($\psi$, **Unachievable**)
  {
  if ($\psi$ != *Repair*)
  {
  *Execute-individual-operator*(*Repair*, self, **C**, $\{\psi$, cause-of-unachievability,...$\}$)
  /* Repair is explained in detail in Section 4.2. Cause-of-unachievability is only domain-specific
  unachievability condition. This is passed as a parameter to repair. */
  } else {
  *Execute-individual-operator*(*Complete-Failure*, self, **C**, $\{\psi$, cause-of-unachievability,...$\}$)
  /* If Repair is itself unachievable, complete-failure results, as in Section 4.2 */
  }
  }

} /* end procedure execute-individual-operator */

## Appendix B: STEAM Sample Rules

The sample rules described below follow the description of STEAM provided in this article, and essentially help encode the algorithm described in Appendix A. The rules, as with the algorithm in Appendix A, are based on execution of hierarchical operators, or reactive plans. While the sample rules below are described in simplified if-then form, the actual rules are encoded in Soar, and are available as an online Appendix.

    **SAMPLE:RULE:CREATE-COMMUNICATIVE-GOAL-ON-ACHIEVED**
    /* This rule focuses on generating a communicative goal
    if an agent's private state contains a belief that satisfies
    the achievement condition of a team operator $[OP]_\Theta$.
    See section 4.1. */
    IF
    agent $\nu$i's private state contains a fact F
    AND
    fact F matches an achievement condition AC
    of a team operator $[OP]_\Theta$
    AND
    fact F is not currently mutually believed
    AND
    a communicative goal for F is not already generated
    THEN

create possible communicative goal CG to communicate fact F to team
$\Theta$ to terminate $[OP]_\Theta$.

### SAMPLE:RULE:CREATE-COMMUNICATIVE-GOAL-ON-UNACHIEVABLE
/* This rule is similar to the one above. */
IF
agent $\nu i$'s private state contains a fact F
AND
fact F matches an unachievability condition UC
of a team operator $[OP]_\Theta$
AND
fact F is not currently mutually believed
AND
a communicative goal for F is not already generated
THEN
create possible communicative goal CG to communicate fact F to team
$\Theta$ to terminate $[OP]_\Theta$.

### SAMPLE:RULE:ESTIMATE-VALUE-FOR-NON-COMMUNICATION
/* This rule estimates $\tau * C_{mf}$ for non-communication
given a communicative goal, using the formula from
Section 5.*/
IF
CG is a possible communicative goal to communicate fact F to team
$\Theta$ to terminate $[OP]_\Theta$
AND
$C_{mt}$ is estimated **high**
AND
$\tau$ is estimated **low**
THEN
Estimated value of non-communication is **medium**.

### SAMPLE:RULE:DECISION-ON-COMMUNICATION
/* This rule makes the communication decision using the formula * with $\tau * C_{mt}$ and Cc
from Section 5.*/
IF
CG is a possible communicative goal to communicate fact F to team
$\Theta$ to terminate $[OP]_\Theta$
AND
Estimated value of non-communication for CG is **medium**
AND
Estimated value of communication for CG is **low**
THEN
post CG as a communicative goal to communicate fact F to team
$\Theta$ to terminate $[OP]_\Theta$

### SAMPLE:RULE:MONITOR-UNACHIEVABILITY:AND-COMBINATION
/* This rule checks for unachievability of role-monitoring
constraints involving an AND-combination. See section 4.2.
/

IF
A current joint intention [OP]$_\Theta$ involves an AND-combination
AND
$\nu$i is a member performing role to execute sub-operator op
AND
no other member $\nu$j is also performing role to execute sub-operator op
AND
$\nu$i cannot perform role
THEN
Current joint intention [OP]$_\Theta$ is unachievable, due to a critical role failure
of $\nu$i in performing op

# References

Barbuceanu, M., & Fox, M. (1996). The architecture of an agent building shell. In Wooldridge, M., Muller, J., & Tambe, M. (Eds.), *Intelligent Agents, Volume II: Lecture Notes in Artificial Intelligence 1037*. Springer-Verlag, Heidelberg, Germany.

Calder, R. B., Smith, J. E., Courtemanche, A. J., Mar, J. M. F., & Ceranowicz, A. Z. (1993). Modsaf behavior simulation and control. In *Proceedings of the Conference on Computer Generated Forces and Behavioral Representation*.

Chu-Carroll, J., & Carberry, S. (1996). Conflict detection and resolution in collaborative planning. In Wooldridge, M., Muller, J., & Tambe, M. (Eds.), *Intelligent Agents, Volume II: Lecture Notes in Artificial Intelligence 1037*. Springer-Verlag, Heidelberg, Germany.

Cohen, P. R., & Levesque, H. J. (1991a). Confirmation and joint action. In *Proceedings of the International Joint Conference on Artificial Intelligence*.

Cohen, P. R., & Levesque, H. J. (1991b). Teamwork. *Nous, 35*.

Coradeschi, S. (1997). A decision mechanism for reactive and coordinated agents. Tech. rep. 615, Linkoping University. (Licentiate Thesis).

Decker, K., & Lesser, V. (1995). Designing a family of coordination algorithms. In *Proceedings of the International Conference on Multi-Agent Systems*.

Durfee, E., & Lesser, V. (1991). Partial global planning: a coordination framework for distributed planning. *IEEE transactions on Systems, Man and Cybernetics, 21*(5).

Firby, J. (1987). An investigation into reactive planning in complex domains. In *Proceedings of the National Conference on Artificial Intelligence (AAAI)*.

Gmytrasiewicz, P. J., Durfee, E. H., & Wehe, D. K. (1991). A decision theoretic approach to co-ordinating multi-agent interactions. In *Proceedings of International Joint Conference on Artificial Intelligence*.

Grosz, B. (1996). Collaborating systems. *AI magazine, 17*(2).

Grosz, B., & Kraus, S. (1996). Collaborative plans for complex group actions. *Artificial Intelligence, 86*, 269–358.

Grosz, B. J., & Sidner, C. L. (1990). Plans for discourse. In Cohen, P. R., Morgan, J., & Pollack, M. (Eds.), *Intentions in Communication*, pp. 417–445. MIT Press, Cambridge, MA.

Halpern, J. Y., & Moses, Y. (1990). Knowledge and common knowledge in a distributed environment. *Journal of the ACM, 37*(3), 549–587.

Hayes-Roth, B., Brownston, L., & Gen, R. V. (1995). Multiagent collaboration in directed improvisation. In *Proceedings of the International Conference on Multi-Agent Systems (ICMAS-95)*.

Hill, R., Chen, J., Gratch, J., Rosenbloom, P., & Tambe, M. (1997). Intelligent agents for the synthetic battlefield: a company of rotary wing aircraft. In *Proceedings of the Innovative Applications of Artificial Intelligence*.

Jennings, N. (1994). Commitments and conventions: the foundation of coordination in multi-agent systems. *The Knowledge Engineering Review, 8*.

Jennings, N. (1995). Controlling cooperative problem solving in industrial multi-agent systems using joint intentions. *Artificial Intelligence, 75*.

Kaminka, G. A., & Tambe, M. (1997). Social comparison for failure monitoring and recovery in multi-agent settings. In *Proceedings of the National Conference on Artificial Intelligence*, p. (Student abstract).

Kinny, D., Ljungberg, M., Rao, A., Sonenberg, E., Tidhard, G., & Werner, E. (1992). Planned team activity. In Castelfranchi, C., & Werner, E. (Eds.), *Artificial Social Systems, Lecture notes in AI 830*. Springer, NY.

Kitano, H., Asada, M., Kuniyoshi, Y., Noda, I., & Osawa, E. (1995). Robocup: The robot world cup initiative. In *Proceedings of IJCAI-95 Workshop on Entertainment and AI/Alife*.

Kitano, H., Tambe, M., Stone, P., Veloso, M., Noda, I., Osawa, E., & Asada, M. (1997). The robocup synthetic agents' challenge. In *Proceedings of the International Joint Conference on Artificial Intelligence (IJCAI)*.

Laird, J. E., Jones, R. M., & Nielsen, P. E. (1994). Coordinated behavior of computer generated forces in tacair-soar. In *Proceedings of the Fourth Conference on Computer Generated Forces and Behavioral Representation*. Orlando, Florida: Institute for Simulation and Training, University of Central Florida.

Levesque, H. J., Cohen, P. R., & Nunes, J. (1990). On acting together. In *Proceedings of the National Conference on Artificial Intelligence*. Menlo Park, Calif.: AAAI press.

Lochbaum, K. E. (1994). *Using collaborative plans to model the intentional structure of discourse*. Ph.D. thesis, Harvard University.

Mitchell, T. M., Keller, R. M., & Kedar-Cabelli, S. T. (1986). Explanation-based generalization: A unifying view. *Machine Learning, 1*(1), 47–80.

Newell, A. (1990). *Unified Theories of Cognition.* Harvard Univ. Press, Cambridge, Mass.

Pimentel, K., & Teixeira, K. (1994). *Virtual reality: Through the new looking glass.* Windcrest/McGraw-Hill, Blue Ridge Summit, PA.

Pollack, M. (1992). The uses of plans. *Artificial Intelligence, 57,* 43–68.

Rajput, S., & Karr, C. R. (1995). Cooperative behavior in modsaf. Tech. rep. IST-CR-95-35, Institute for simulation and training, University of Central Florida.

Rao, A. S., Lucas, A., Morley, D., Selvestrel, M., & Murray, G. (1993). Agent-oriented architecture for air-combat simulation. Tech. rep. Technical Note 42, The Australian Artificial Intelligence Institute.

Reilly, W. S. (1996). *Believable Emotional and Social Agents.* Ph.D. thesis, School of Computer Science, Carnegie Mellon University.

Rich, C., & Sidner, C. (1997). COLLAGEN: When agents collaborate with people. In *Proceedings of the International Conference on Autonomous Agents (Agents'97).*

Rosenbloom, P. S., Laird, J. E., Newell, A., , & McCarl, R. (1991). A preliminary analysis of the soar architecture as a basis for general intelligence. *Artificial Intelligence, 47*(1-3), 289–325.

Sen, S. (1996). *Proceedings of the Spring Symposium on Adaptation, Coevolution and Learning.* American Association for Artificial Intelligence, Menlo Park, CA.

Sidner, C. (1994). An artificial discourse language for collaborative negotiation. In *Proceedings of the National Conference on Artificial Intelligence (AAAI).*

Smith, I., & Cohen, P. (1996). Towards semantics for an agent communication language based on speech acts. In *Proceedings of the National Conference on Artificial Intelligence (AAAI).*

Sonenberg, E., Tidhard, G., Werner, E., Kinny, D., Ljungberg, M., & Rao, A. (1994). Planned team activity. Tech. rep. 26, Australian AI Institute.

Stone, P., & Veloso, M. (1996). Towards collaborative and adversarial learning: a case study in robotic soccer. In Sen, S. (Ed.), *AAAI Spring Symposium on Adaptation, Coevolution and Learning in multi-agent systems.*

Tambe, M. (1995). Recursive agent and agent-group tracking in a real-time dynamic environment. In *Proceedings of the International Conference on Multi-agent systems (ICMAS).*

Tambe, M. (1996). Tracking dynamic team activity. In *Proceedings of the National Conference on Artificial Intelligence (AAAI).*

Tambe, M. (1997a). Agent architectures for flexible, practical teamwork. In *Proceedings of the National Conference on Artificial Intelligence (AAAI)*.

Tambe, M. (1997b). Implementing agent teams in dynamic multi-agent environments. *Applied Artificial Intelligence*. (to appear).

Tambe, M., Johnson, W. L., Jones, R., Koss, F., Laird, J. E., Rosenbloom, P. S., & Schwamb, K. (1995). Intelligent agents for interactive simulation environments. *AI Magazine*, *16*(1).

Tambe, M., & Rosenbloom, P. S. (1995). RESC: An approach for real-time, dynamic agent tracking. In *Proceedings of the International Joint Conference on Artificial Intelligence (IJCAI)*.

Tambe, M., Schwamb, K., & Rosenbloom, P. S. (1995). Building intelligent pilots for simulated rotary wing aircraft. In *Proceedings of the Fifth Conference on Computer Generated Forces and Behavioral Representation*.

Tidhar, G., Selvestrel, M., & Heinze, C. (1995). Modeling teams and team tactics in whole air mission modeling. Tech. rep. Technical Note 60, The Australian Artificial Intelligence Institute.

Williamson, M., Sycara, K., & Decker, K. (1996). Executing decision-theoretic plans in multi-agent environments. In *Proceedings of the AAAI Fall Symposium on Plan Execution: Problems and Issues*.